\documentclass{article}

\usepackage{microtype}
\usepackage{graphicx}
\usepackage{subfigure}
\usepackage{booktabs}
\usepackage{hyperref}
\usepackage{wrapfig}
\usepackage{float}

\usepackage{booktabs}
\usepackage{siunitx}
\usepackage{adjustbox}

\PassOptionsToPackage{numbers, compress}{natbib}
\usepackage[preprint]{neurips_2026}

\usepackage{amsmath}
\usepackage{amssymb}
\usepackage{mathtools}
\usepackage{amsthm}

\usepackage{relsize}
\usepackage{array}

\usepackage[capitalize,noabbrev]{cleveref}

\theoremstyle{plain}

\theoremstyle{definition}

\theoremstyle{remark}

\usepackage[textsize=tiny]{todonotes}

\usepackage{amsmath,amsfonts,bm}

\def\eqref#1{equation~\ref{#1}}
\def\1{\bm{1}}

\DeclareMathAlphabet{\mathsfit}{\encodingdefault}{\sfdefault}{m}{sl}
\SetMathAlphabet{\mathsfit}{bold}{\encodingdefault}{\sfdefault}{bx}{n}

\usepackage{hyperref}
\usepackage{url}
\usepackage{xspace}
\usepackage{graphicx}
\usepackage{booktabs}
\usepackage{multirow}
\usepackage{multicol}
\usepackage{wrapfig}
\usepackage{adjustbox}
\usepackage{lipsum}

\definecolor{borgogna}{RGB}{159, 29, 53}

\newcommand{\implname}{\textsc{Dream-Cubed}\xspace}

\author{
  Tim Merino\thanks{This work was performed as an internship project at Sakana AI}\\
  New York University \\
  \texttt{tm3477@nyu.edu} \\
  \And
  Sam Earle$^*$\\
  New York University\\
  \texttt{se2181@nyu.edu} \\
  \And
  Ryunosuke Iwai \\
  Sakana AI\\
  \texttt{ryunosukeiwai@sakana.ai} \\
  \And
  Julian Togelius \\
  New York University\\
  \texttt{jt125@nyu.edu} \\
  \And
  Edoardo Cetin \\
  Sakana AI\\
  \texttt{edo@sakana.ai} \\
}
\begin{document}

\title{Dream Cubed: Controllable Generative Modeling in Minecraft by Training on Billions of Cubes}
\maketitle

\begin{abstract}
We introduce Dream-Cubed, a large-scale dataset of Minecraft worlds at voxel resolution, and a family of models using cubes as powerful compositional units for efficient generation of interactive 3D environments. Dream-Cubed comprises tens of billions of tokens from a carefully curated mixture of procedural biome terrain and high-quality human-authored maps. We use this dataset to conduct the first large-scale study of 3D diffusion models for voxel generation, analyzing discrete and continuous diffusion formulations, data compositions, and architectural design choices. Our models operate directly in the space of blocks, enabling efficient and semantically grounded generation while supporting interactive user workflows such as inpainting and outpainting from user-authored blocks. To quantitatively evaluate our models, we adapt the FID metric to assess semantic differences between real and generated world renderings, and validate generation quality through a human preference study. We release the full dataset, code, and all our pretrained models, which we hope will provide a foundation for future research in efficient generative modeling for structured, interactive 3D environments.
\end{abstract}

\begin{figure}[!h]
    \centering
    \includegraphics[width=0.56\linewidth]{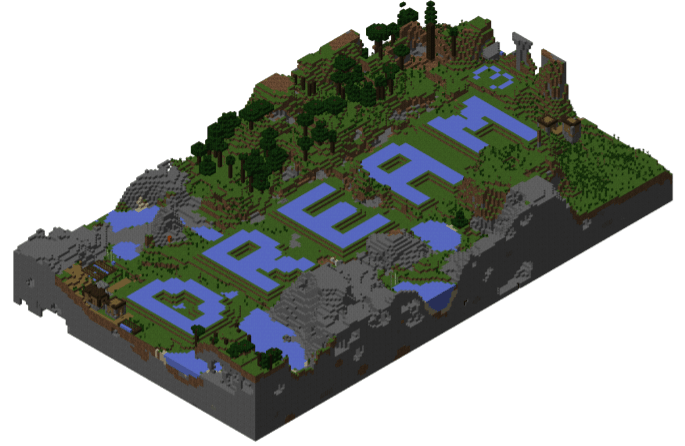}
    \caption{A Minecraft world generated by our block-level diffusion model using user-authored block constraints. The text ``$\text{Dream}^3$'' serves as fixed block tokens during generation. No architectural modification or additional training is required to support this conditioning.}
    \label{fig:logo_figure}
\end{figure}
\section{Introduction}

\label{sec1:intro}
% purpose 
The field of Generative AI has seen incredible progress in image, video, and text generation~\citep{dalle2, gpt4, wan2025wan}. Despite the meteoric rise in both popularity and capability of these models, there has been comparatively little attention paid to interactive 3D environments. This domain encompasses a vast range of discrete, compositional, editable content that underlie the virtual worlds millions interact with in popular video games.

% relevance + dragon
We study controllable generation in Minecraft: the best-selling video game to date. The Minecraft world is represented as a 3D voxel grid, where each block is a semantically meaningful unit that specifies visual appearance and gameplay behavior\footnote{We use the terms \textit{voxels}, \textit{cubes}, and \textit{blocks} interchangeably to denote Minecraft's characterizing block-shaped compositional units.}. As the most popular open-world game, Minecraft has attracted an extensive body of prior research, serving as a challenging, open-ended benchmark for game-playing agents and a testbed for learning complex rule-based transition dynamics for world modeling \cite{dreamer, fan2022minedojo, wang2023voyager, long2024teamcraft, decart2024oasis, savva2026solaris}. However, despite its tremendous value in AI research, little work has focused on block-level world generation, largely due to a lack of large-scale, well-defined datasets and benchmarks that have historically catalyzed progress for the development of generative models~\citep{penntreebank, imagenet, openwebtext}.

To this end, we introduce Dream-Cubed: a first-of-its-kind large-scale dataset of Minecraft terrain at block resolution. The dataset is composed of millions of $32^3$-block chunks that capture a rich snapshot of an open-ended game world, encompassing winding rivers, towering snow-capped mountains, underground cave systems, and structured settlements. Beyond naturally occurring terrain, our data also includes large-scale human-authored maps from professional Minecraft builders, which we share with the authors' consent. For scraping and curating our data, we leverage multiple extraction pipelines that we design to ensure high coverage of semantically meaningful voxel structures with complex geometries. Overall, spanning hundreds of billions of cubes, our dataset pushes the limits of structural complexity and creative diversity, providing a foundation for large-scale generative modeling. 

We use Dream-cubed to conduct the first large-scale study of 3D diffusion models for biome-conditioned voxel generation. Our models diffuse directly in the space of blocks, serving as semantically rich ``tokens'' of abstraction that avoid the complexities of pixel-space modeling while retaining full generation expressivity within Minecraft. We evaluate discrete and continuous diffusion objectives such as MD4 \cite{md4} and DDPM~\citep{karras2022edm, ddpm} to train powerful diffusion transformers~\citep{peebles2023scalable}, tuning the whole pipeline for efficient 3D generation. We study a set of design choices in data composition to architecture optimization, providing numerous model comparisons to facilitate future research. To evaluate our models, we adapt the Frechét Inception Distance (FID)~\citep{heusel2018fid} for assessing semantic differences between real and generated world renderings, and validate our results by running a human preference study.

At inference time, we show that our models enable a new class of efficient, interactive, and controllable generation workflows. Analogous to recent advances in image generation, our models support flexible editing operations such as inpainting, outpainting, and biome blending, allowing users to author partial voxel structures and generate coherent completions at scale (e.g., Figure~\ref{fig:logo_figure}). Acting directly in the space of blocks rather than pixels ensures generated outputs are fully playable within the Minecraft game environment and enables efficient large-scale synthesis, as worlds can be extended, modified, or recomposed without requiring expensive rendering or post-processing steps. We release the full dataset, code, and all our pretrained models with a permissive open-source license to support and facilitate future research in generative modeling for interactive 3D environments \footnote{Code, pretrained model weights, and data are available at https://github.com/SakanaAI/DreamCubed}.

In summary, our main contributions are as follows:
\begin{itemize}
    \item We release \implname, a large-scale curated dataset of over 2 million Minecraft chunks spanning tens of billions of cubes from diverse biome-labeled terrain and high-quality human-authored maps.
    \item We present a comparison of training discrete and continuous 3D diffusion models, analyzing the impact of different data compositions and architectural design choices on voxel-level generation.
    \item We adapt the FID metric to evaluate model quality, validate our results with a human-preference study, and demonstrate how our models naturally enable a new class of efficient, interactive, and controllable generation workflows for interactive 3D environments.
\end{itemize}

\section{Dataset}
\paragraph{Overview}
Our dataset is organized into two main subsets. The core dataset consists of $1,667,781$ chunks of procedurally generated Minecraft terrain, including natural overworld, underground caves, and villages, which we label using fifteen distinct biome classes and represent using a vocabulary of $V = 117$ block types. The supplementary dataset includes human-authored data, consisting of $358,762$ chunks from maps built by professional community builders, which expand the vocabulary to $V = 177$ block types and introduce six additional map-specific labels. We train separate models either on the core or human-augmented dataset to investigate generative capability for increasingly complex structured content.

\paragraph{Data representation}
We study conditional generation of Minecraft worlds at the block level. Each sample in our dataset is a tensor $\mathbf{x}$ of shape $32 \times 32 \times 32$, where $x_{i,j,k}$ is an integer block ID in $\{1, \dots, V\}$.  Each discrete block ID acts as a separate input token to our generative models. We condition our models on a chunk-level biome label $\mathbf{y}$. As Minecraft exposes a biome label (e.g. ``desert'') for each voxel in the game world, we assign a ``global'' biome label to the chunk equal to the biome label with the highest frequency across its voxels. We summarize the vocabulary construction, including block state metadata handling and compression of the human-authored block palette in Appendix \ref{apdx:vocab}.

\subsection{Data Composition}
We construct \implname using four targeted data collection pipelines, each capturing distinct categories of Minecraft content:
\begin{enumerate}
    \item Natural biome chunks sampled from the surface of the world
    \item Cave chunks sampled from the game's underworld
    \item Village chunks sampled densely from procedurally generated villages
    \item Human chunks sampled from user-generated builds 
\end{enumerate}

\begin{wraptable}{r}{0.35\textwidth}
\centering
\vspace{-8mm}
\caption{Chunk counts for each natural biome}
\vspace{4mm}
\label{tab:biome_chunk_count}
\begin{tabular}{lr}
Biome & Total Chunks \\
\midrule
Beaches & 115,058 \\
Birch forest & 126,148 \\
Cave & 146,000 \\
Desert & 143,208 \\
Extreme hills & 93,030 \\
Forest & 142,942 \\
Ice & 99,579 \\
Jungle & 77,560 \\
Ocean & 234,157 \\
Plains & 98,017 \\
River & 69,192 \\
Savanna & 94,395 \\
Swampland & 118,753 \\
Taiga & 109,742 \\
Village & 178,271 \\
\bottomrule
\end{tabular}
\vspace{-8mm}
\end{wraptable}

We motivate each category below. Full implementation details, including filtering criteria and biome label mappings, are provided in Appendix \ref{apdx:data_collection}. Table \ref{tab:biome_chunk_count} provides a final per-biome chunk count for each biome in the core \implname dataset.

\paragraph{Natural overworld} Minecraft worlds are generated using a biome system, partitioning the overworld terrain into regions based on a random seed. Each biome specifies block composition and characteristic terrain features of that region; deserts contain sand and cacti, while oceans have deep expanses of water. We manually partition the dozens of Minecraft biomes into a representative set of 13 biome classes, which make up the majority of our dataset. We scrape this data from procedurally generated Minecraft worlds, collecting $1{,}521{,}781$ biome-labeled natural biome chunks.

\paragraph{Underworld collection} Underground caves, ravines, and caverns are an iconic part of Minecraft, but they are not represented as a dedicated biome category in the base game. Given their importance in the game and notorious semantic complexity, we introduce an additional ``cave'' label and construct a targeted collection pipeline to comprehensively cover underground constructions, yielding a total of 146,000 cave chunks

\paragraph{Villages collection} 
NPC villages are unique in Minecraft, as they contain several different complex structures such as houses and farms. These are also a relatively rare occurrence in the Minecraft world, and are not inherently represented with a dedicated biome category. Thus, we introduce an additional ``village'' label and design a targeted sampling strategy to ensure adequate representation, yielding 178,271 curated village chunks.

\paragraph{Human-designed environments collection} We complement our procedurally generated terrain with human-designed maps sourced with permission from popular Minecraft creators\footnote{\texttt{https://www.planetminecraft.com/member/timtenth\_buildings/}}. We collect chunks from six maps, targeting high-quality, large maps that contain many structures with a consistent style. We treat each map as a new biome, using the map's name as a biome label for all collected chunks. In total, we collect 358,762 human-authored chunks across six maps.

\label{sec2:dataset}
\section{Method}
% \subsection{Modeling formulations.}
We compare two families of diffusion models for biome-conditioned chunk generation. Given that Minecraft blocks are categorical tokens, a natural question is whether a generative model should operate in discrete or continuous space. To investigate this, we implement both discrete and continuous diffusion\footnote{We use ``continuous'' to refer to the continuous-valued representation of tokens, not continuous-time diffusion.} with an identical 280M-parameter 3D Diffusion Transformer backbone. Both model types generate a single $32\times 32\times 32$ chunk conditioned on a biome label $y$ and timestep $t$.

(i) \textbf{Discrete masked diffusion (MD4):} Each voxel is treated as a categorical token over a vocabulary of $V$ block types. The model learns an iterative unmasking model, progressively replacing masked tokens with predicted block types.

(ii) \textbf{DDPM in embedding space:} Each voxel is mapped to a fixed $d$-dimensional vector.
A discrete-time DDPM model with $v$-prediction operates in the resulting continuous tensor space, and a decoding step recovers discrete block types.

\subsection{Shared Architecture}
Both formulations use the same 3D Diffusion Transformer (DiT) architecture design, with approximately 280M parameters designed to strike a balance between expressivity and fast generation. Each network consists of 25 transformer blocks, with a hidden dimension of $768$, and $8$ attention heads. The two formulations differ in their network's input and output projections, which handle discrete tokens in the MD4 case, or continuous embeddings in the DDPM case.

Each input $32 \times 32 \times 32$ chunk is partitioned into non-overlapping voxel patches using a 3D convolution with kernel size and stride equal to the patch size $p$, yielding a grid of $(32 / p)^3$ patch tokens. Patch positions are encoded with fixed 3D sine--cosine embeddings, while timesteps are encoded via sinusoidal embeddings followed by a multi-layer perceptron, and biome labels via a learned class embedding followed by a separate MLP. The timestep and class embeddings are summed and projected into the transformer's hidden dimension. Conditioning is injected into every transformer block through adaptive layer normalization (AdaLN) modulation and gating, rather than through prefix tokens or cross-attention. This design is agnostic to the diffusion formulation, allowing the same backbone to serve both MD4 and DDPM.

\subsection{Discrete masked diffusion (MD4)}
\label{sec3:implementation:md4}
We implement MD4 masked diffusion for discrete 3D voxel grids \cite{md4}. MD4 augments the token vocabulary with an additional \texttt{[MASK]} token. The forward process progressively masks tokens, while the reverse process iteratively unmasks them.

\paragraph{Forward process \& training objective}
Given a clean chunk $x_0$, we sample a timestep $t\sim \mathrm{Uniform}(0,1)$ and independently mask each voxel with probability $p_{\text{mask}}(t) = \sin\!\left(\frac{\pi t}{2}\right)$, following \cite{md4}. During training, the model receives a corrupted chunk $x_t$ as a one-hot tensor over $V+1$ classes (including the \texttt{[MASK]} state), along with the current timestep $t$ and biome label $y$. The network outputs logits over $V$ real block labels for each voxel. We train with cross-entropy on masked positions only, and we apply a time-dependent weighting following MD4.

\paragraph{Sampling}
Generation begins from a fully-masked chunk $x_T$, and proceeds backwards over a discrete set of timesteps from $t=1$ to $t =0$. At each step,an unmask probability $p_{\text{unmask}}$ from $t_i$ to $t_{i-1}$ is used to randomly select tokens to unmask, replacing them with token predictions via multinomial sampling from the softmax of the model's output logits. This continues until all tokens are unmasked.

\subsection{Continuous diffusion in embedding space}
\label{sec3:implementation:ddpm}
As a continuous alternative, we implement a discrete-time DDPM on voxel embeddings, obtained via a pretrained text-embedding model.

\paragraph{Continuous block embeddings}
Continuous diffusion requires a dense vector representation for each block type. We leverage the fact that Minecraft block types have semantically meaningful names (e.g., dirt", ``sand") and obtain fixed $16$-dimensional embeddings by encoding each block's name string using OpenAI's \texttt{text-embedding-3-small} model \cite{openai2024embeddings}. This gives a precomputed embedding table $E \in \mathbb{R}^{V \times 16}$ that remains frozen during training, and provides a semantic prior without introducing additional training.

\paragraph{Forward process \& training objective}
We use a cosine noise schedule over 1,000 discrete steps \cite{nichol2021improved}. At training time, we sample a random timestep $t$ and corrupt the clean embedded chunk $x_0$ by mixing it with Gaussian noise:
\begin{equation}
x_t = a_t\,x_0 + b_t\,\epsilon,\qquad \epsilon\sim\mathcal{N}(0,I),
\end{equation}
where $(a_t,b_t)$ are schedule-defined coefficients with $a_t^2 + b_t^2 = 1$. We use the velocity ($v$) parameterization \cite{salimans2022progressive}, and train the network with an MSE loss to predict $v_t$ from $(x_t,t, y)$.

\paragraph{Sampling.}
Generation begins from pure Gaussian noise $x_T \sim \mathcal{N}(0, I)$ and backwards over a discrete set of timesteps from $t=1$ to $t =0$. At each reverse step, we convert the predicted velocity $\hat{v}$ to a denoised estimate $\hat{x}_0 = a_t x_t - b_t \hat{v}$, then compute the next (less noisy) state $x_{t-1}$ via the standard DDPM posterior.
Once sampling is complete, the final continuous tensor $\hat{x}_0$ must be decoded back to discrete block IDs. We perform per-voxel nearest-neighbor lookup in our embedding table under L2 distance, recovering the closest block ID for each denoised voxel embedding.

\subsection{Patch size sensitivity.}
We find that the two formulations differ substantially in their sensitivity to the patch size hyperparameter. MD4 models generate coherent chunks across a range of patch sizes:  $p = 2$ (4096 patch tokens) and $p = 4$ (512 tokens) both produce high-quality outputs, while $p = 8$ (64 tokens) yields recognizable outputs with noticeable visual artifacts. $p = 1$ (32,768 tokens) proves prohibitively expensive for training. In contrast, DDPM produces high-quality outputs at $p=2$, but fails at $p=4$ under an otherwise identical configuration. Given this, we train MD4 models at patch sizes 2 and 4, and DDPM models at patch size 2.

\label{sec3:implementation}

\section{Quantitative Results}
We evaluate biome-conditioned chunk generation using Fr\'echet Inception Distance (FID), computed on rendered images of generated and real chunks \cite{heusel2018fid}. For each biome, we generate 100 samples and compute FID against a fixed set of chunks from the validation set, evenly distributed by biome. We compute a reference FID for each biome using 100 held-out chunks, and report the adjusted per-biome FID for model comparisons (generated FID minus reference FID).

FID is an image-domain metric, and cannot directly measure 3D voxel quality---a completely  ``hollow'' map (i.e., only surface voxels) may render well despite poorly matching the true data distribution. 
However, chunk renders encode salient details for our domain, such as block composition through texture and color statistics, and geometric structure via silhouette. Lacking established metrics for generation quality in this domain, FID presents a practical proxy metric. We compliment this with block distribution metrics in Appendix \ref{tab:voxel_coverage_non_air} and \ref{tab:voxel_js_non_air}, and investigate alignment with human judgment in section \ref{sec:user_study}. 

\begin{figure}[t]
  \centering
  \includegraphics[width=\linewidth]{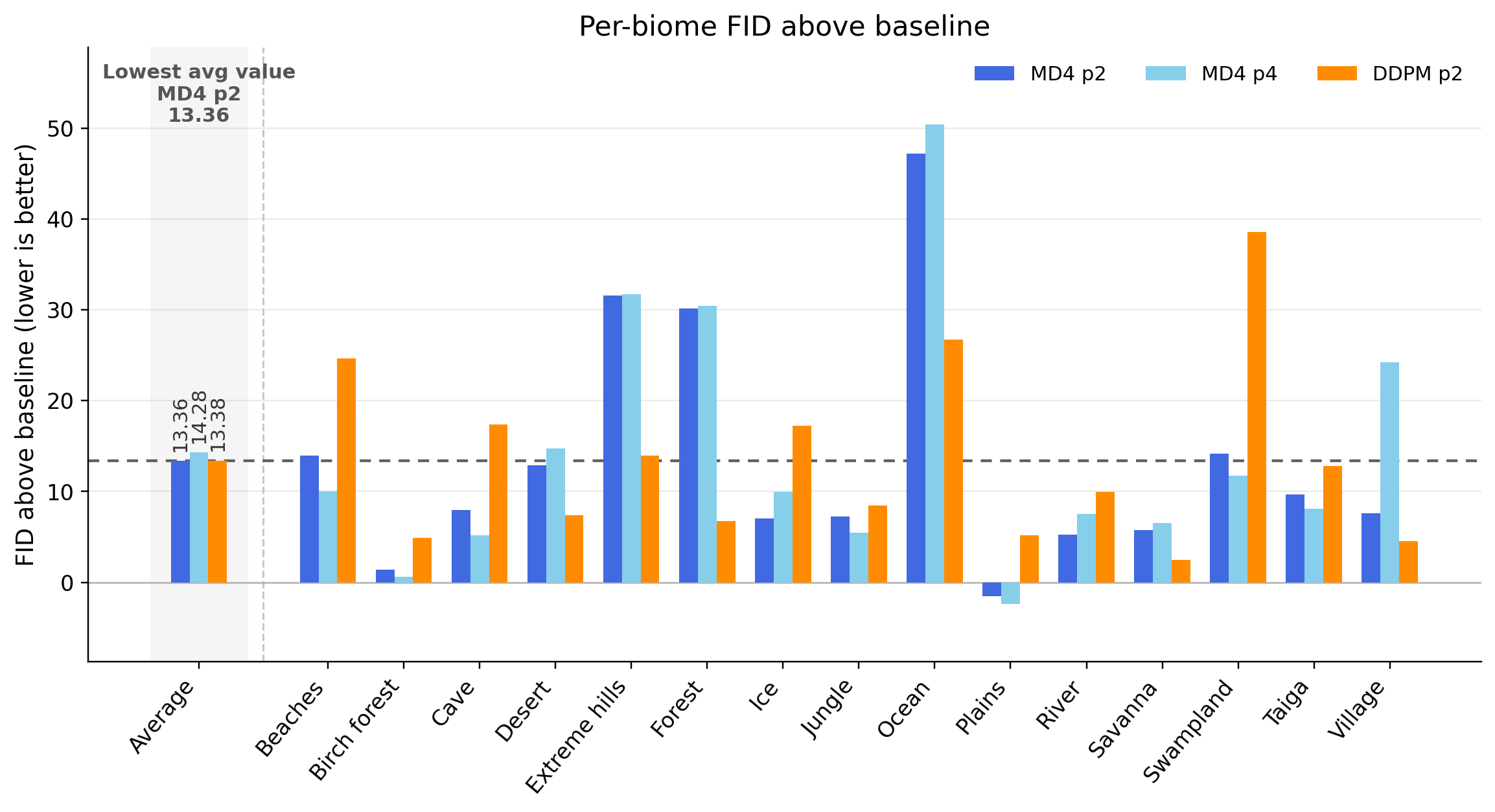}
  \caption{Adjusted per-biome FID for MD4 (patch 2) and MD4 (patch 4) and DDPM (patch 2) models. Lower is better.}
  \label{fig:md4_vs_md4_vs_ddpm}
\end{figure}
\paragraph{MD4 vs.\ DDPM.}
Figure \ref{fig:md4_vs_md4_vs_ddpm} presents a direct comparison between discrete and continuous diffusion under identical conditions. Both models are trained with the same natural biome dataset, with patch size 2 and an identical 280M parameter DiT backbone. The two models achieve almost identical performance on average (59.26 vs 59.29). Despite differences in per-biome FID, neither model dominates overall: MD4 achieves the lower FID in 9 biomes, while DDPM wins in the other 6.

\paragraph{Patch size}
MD4 models trained with patch size 2 and 4 achieve similar average FID (59.26 vs 60.64), 
though the patch 2 model achieves a notably lower FID on the village biome (61.65 vs 78.27). While patch 4 models are substantially cheaper to sample from, due to the $8 \times$ shorter sequence length, these results indicate a finer resolution may be preferable when generating structured content. We also observe a meaningful qualitative difference in controllability, where patch 2 models are noticeably more robust to out-of-distribution sampling conditions. For this reason, we use the patch 2 configuration for training on human-authored content (Appendix \ref{apdx:human_Data}) and for controllable generation experiments in section \ref{sec:controllable_generation}.

\paragraph{Data composition}
We investigate how dataset composition affects generation quality by training the same MD4 architecture (patch size 4, ${\approx}$280M parameters) on three dataset variants:

\begin{enumerate}
    \item \textbf{Balanced Dataset:} Equal representation for all 15 biome classes ($\approx$ 67k samples each).
    \item \textbf{Natural Occurrence:} Each biome is represented approximately equal to its occurrence in a randomly generated Minecraft world. We describe this process and the final dataset distribution in Appendix \ref{apdx:natural_occurrence}.
    \item \textbf{Village-boosted:} Village samples doubled, other biomes remain balanced but reduced proportionally.
\end{enumerate}

\begin{figure}[t]
  \centering
  \includegraphics[width=\linewidth]{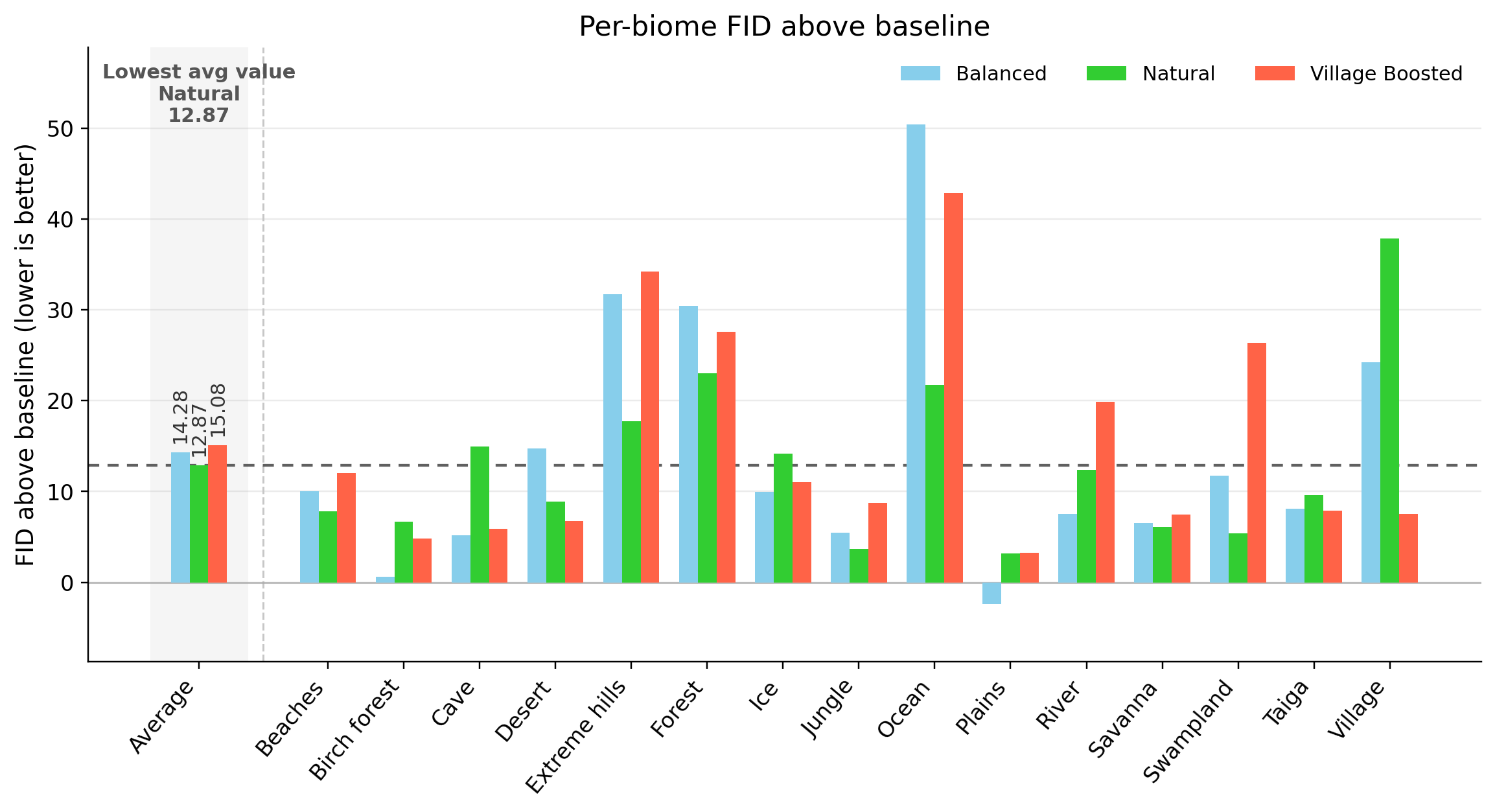}
  \caption{Adjusted per-biome FID comparison between MD4 (patch 4) models across three dataset composition variants. Lower is better.}
  \label{fig:data_balance_fid_fig}
\end{figure}

Figure \ref{fig:data_balance_fid_fig} shows adjusted per-biome FID for each data composition variant. All three datasets achieve similar average FID, with natural occurrence achieving the best. Per-biome FID scores show notable differences arising from the difference in data distribution. The natural occurrence dataset shows large gains in heavily overrepresented biomes, such as ocean (5.3$\times$ the balanced share) and forest ($1.7 \times$), with significant degradation in underrepresented biomes like village and cave. The village boosted dataset shows a large improvement in the village biome, showing that targeted oversampling benefits performance in structured biomes. These results motivate the targeted collection pipelines we introduce in \implname.

\section{Controllable Generation}
\label{sec:controllable_generation}
Because our models operate directly in the block space, and MD4's masking formulation guarantees that unmasked tokens remain fixed throughout sampling, any set of blocks can be imposed as hard constraints during sampling. This property arises by construction in masked diffusion, while the analogous process in DDPM (injecting clean embeddings during sampling) would violate the noise schedule the model was trained on. This enables a spectrum of controllable generation modes --- from local inpainting to large-scale world synthesis --- as a byproduct of training biome-conditional MD4 models. We demonstrate this using qualitative samples from our MD4 (patch 2) model, starting from the simple inpainting case and working up to human-authored block conditions.

\begin{figure*}[!ht]
    \centering
    \includegraphics[width=\linewidth]{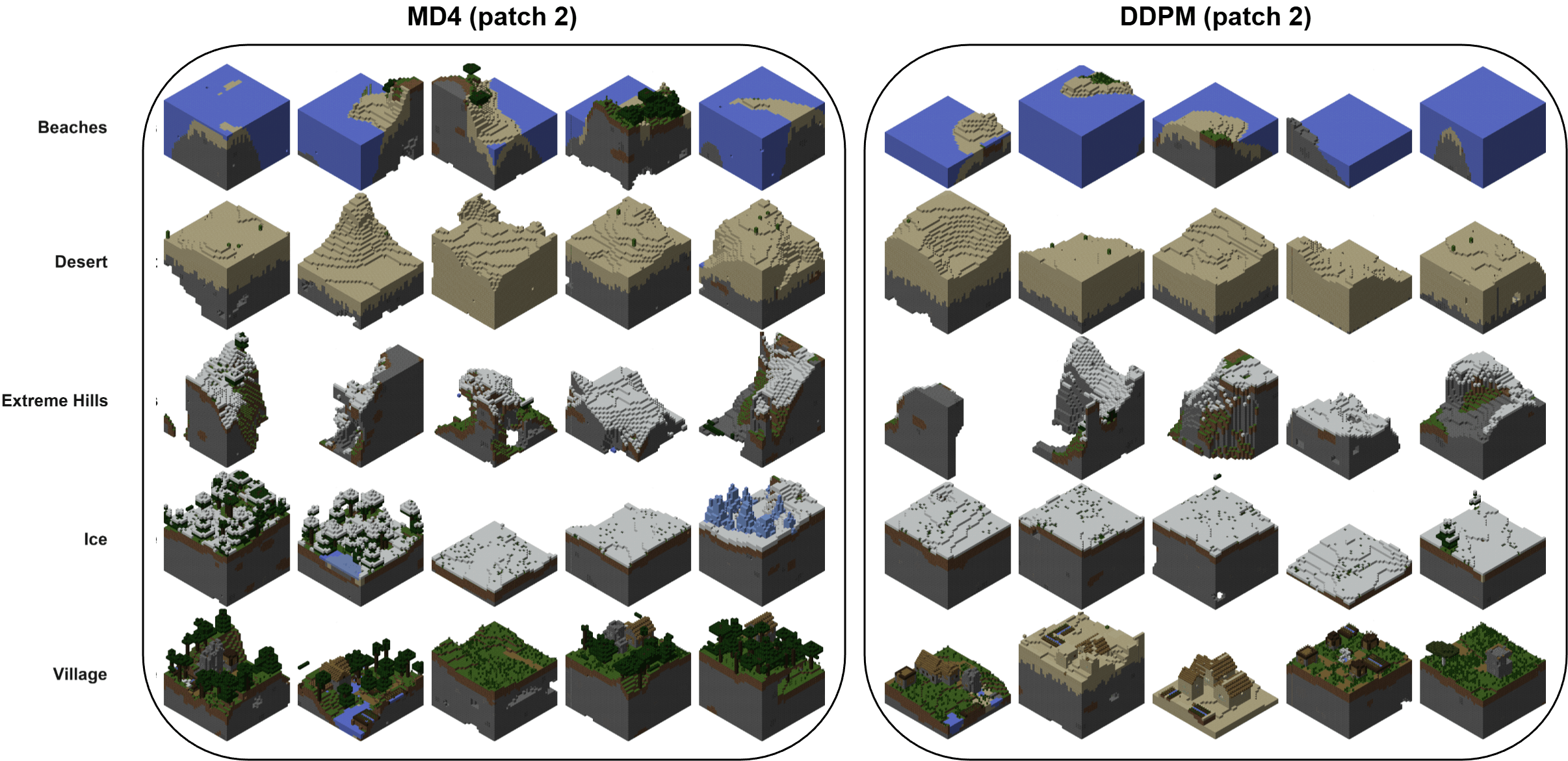}
    \caption{Biome-conditioned samples generated by our MD4 and DDPM models across 5 natural biome classes}
    \label{fig:combined_biome_samples}
    \vspace{8mm}
    \centering
    \includegraphics[width=\linewidth]{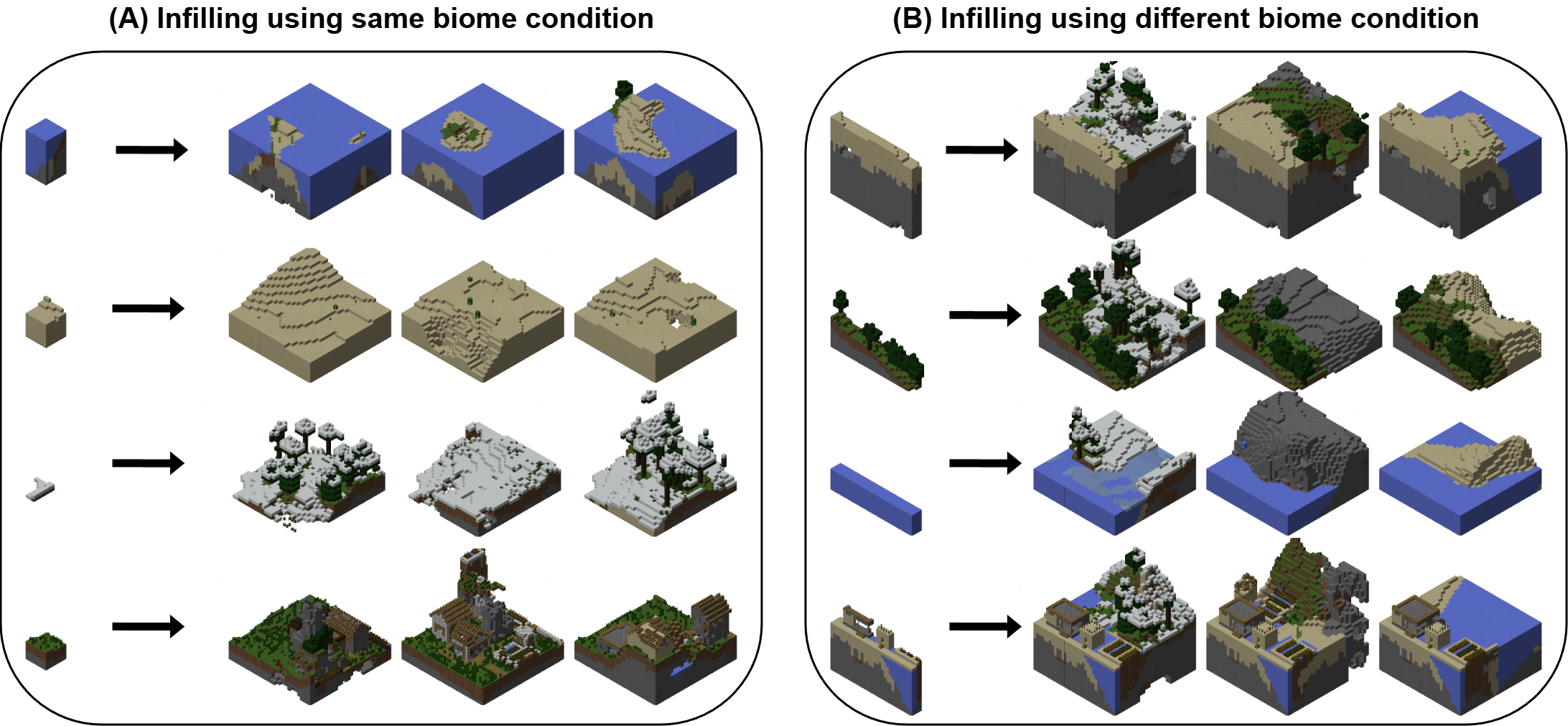}
    \caption{Inpainting from an existing chunk's blocks. (A) Inpainting using the source chunk's biome during sampling. (B) Inpainting using Ice, Extreme Hills, and Beaches biome during sampling. }
    \label{fig:inpaint}
\end{figure*}

\paragraph{Biome-conditioned generation}
Figure \ref{fig:combined_biome_samples} shows randomly selected samples generated by our MD4 and DDPM (patch size 2) models, conditioned on select biome labels. Both models produce chunks that are diverse in terrain features and geometry while adhering to the styles of the condition biome.

\paragraph{Inpainting}

We construct a standard inpainting scenario, where a subset of an existing chunk is fixed, and the remaining region is regenerated using the same biome condition. Figure \ref{fig:inpaint} (A) shows four inpainted chunks conditioned on an $8 \times 32 \times 8$ slice of an existing sample from the validation set. Starting from an identical context, the model produces diverse completions that respect the existing geometry while varying substantially in overall terrain structure.

\paragraph{Biome blending}
We generate completions from a fixed source chunk while providing a different biome label during sampling, to evaluate our model's capability to reconcile conflicting block and biome conditions. Figure \ref{fig:inpaint} B shows inpainted chunks using a fixed $4 \times 32 \times 32$ slice of an existing validation chunk. We generate variants conditioned using the ice, extreme hills, and beaches biomes, demonstrating how our model organically blends between biomes.

\begin{figure*}[ht]
    \centering
    \includegraphics[width=\textwidth]{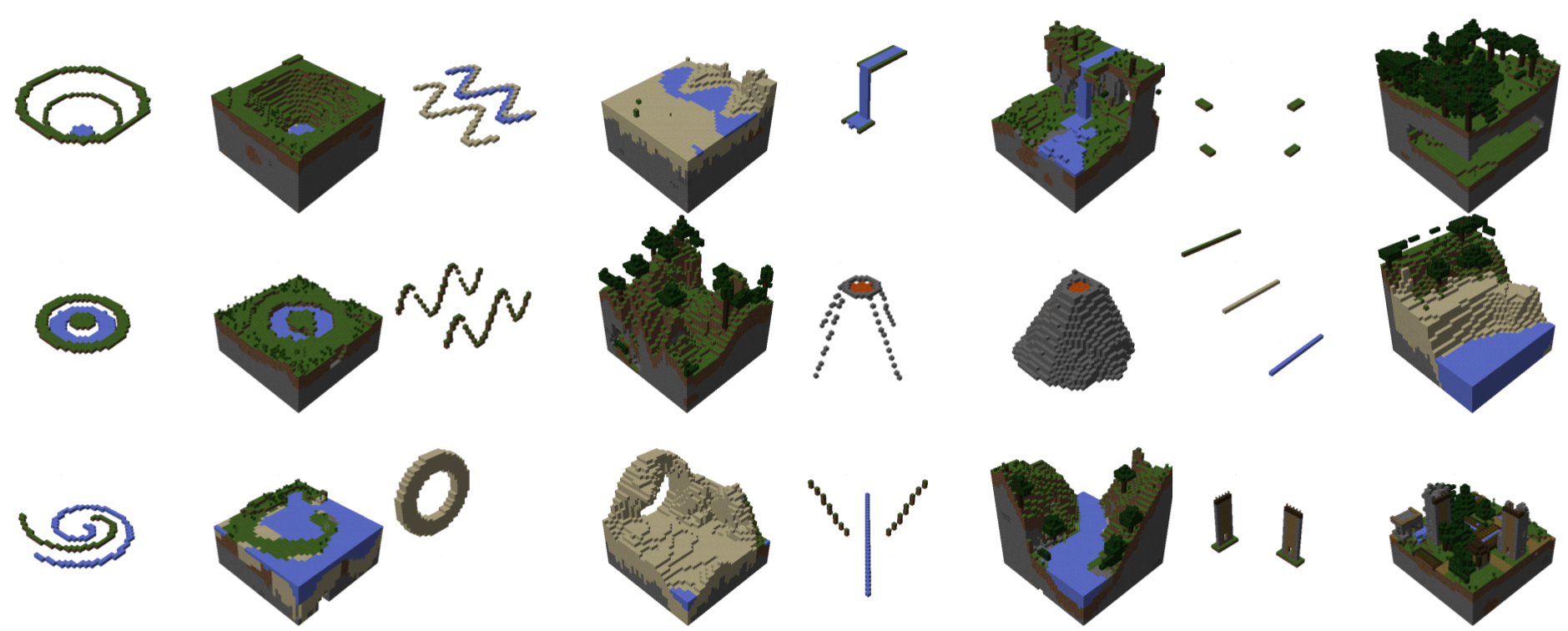}
    \caption{Inpainting using manually authored block patterns as context. Each pair shows the context blocks (left) and the corresponding inpainted chunk (right). Invisible \emph{air} blocks are placed above all manually authored blocks, to ensure the model does not bury the authored pattern underground.}
    \label{fig:manual_context_infill}
\end{figure*}

\paragraph{Block Conditioning} 
We extend inpainting to a more challenging setting: conditioning on arbitrary, user-generated blocks. We manually construct sparse block-level ``prompts'' that are intentionally out of distribution relative to the training data. These include geometric patterns such as sine waves and spirals, as well as more semantically suggestive structures such as volcanoes and waterfalls. 
Figure \ref{fig:manual_context_infill} shows chunks completions using these patterns as fixed context. The model is remarkably robust, generating coherent completions that respect and incorporate the authored block pattern while extending it into plausible surrounding terrain. These examples suggest that the model has learned reusable terrain priors that can generalize beyond the training data distribution.

\begin{figure*}[!ht]
    \centering
    \includegraphics[width=\textwidth]{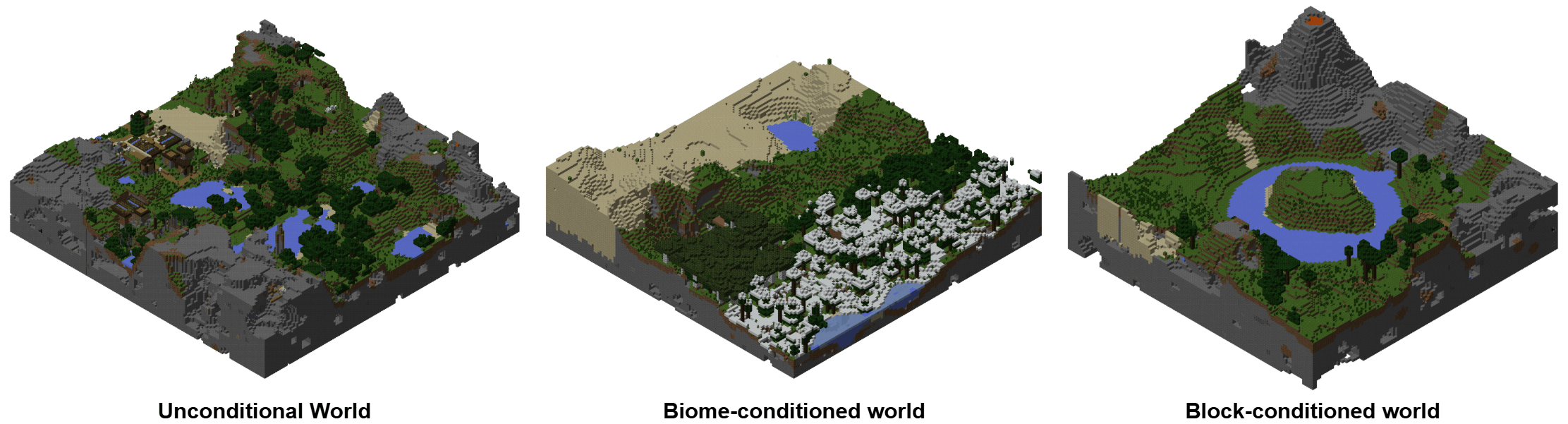}
    \caption{Large-scale ``worlds'' generated via outpainting. Left: A $5\times5\times1$ world generated without any conditioning. Center:  A $4\times4\times1$ world generated with biome conditioning. Right: $4\times4\times1$ world generating using manually authored block contexts, specifying a ring of water and a volcano.}
    \label{fig:combined_world_gen}
\end{figure*}

\paragraph{Generation beyond native resolution}
Training block-level diffusion models on full-scale Minecraft worlds is computational infeasible; the $32^3$ chunk size we use is a necessary compromise to keep sequence lengths tractable. Meaningful applications of controllable generation, from environment generators for game-playing agents to co-creative tools, require coherent generation at scales beyond a single chunk. To bridge this gap, we implement sliding-window outpainting for generation beyond the training resolution. We partition a larger world canvas into overlapping $32^3$ cells, and generate each sequentially using our inpainting procedure. Each cell (after the first) uses overlapping blocks from previously generated cells as fixed context, enforcing consistency across neighboring regions. This procedure naturally defines a spectrum of controllability:

\emph{Unconditional World Generation:} In the simplest case, the model receives only overlapping voxel context from neighboring cells, generating each chunk without a biome condition. Figure~\ref{fig:combined_world_gen} (left) shows that even without biome targets, the model produces coherent multi-chunk terrain containing multiple biome styles and varied large-scale structure. 

\emph{Biome-conditioned world generation:} By assigning each cell a biome label, we specify coarse semantic layouts over large regions. Figure~\ref{fig:combined_world_gen} (center) illustrates spatially varying biome-conditioned generation using rows of ice, forest, plains, and desert biomes.

\emph{User-guided world generation:} Finally, we introduce exact user-authored block constraints into the outpainting procedure, enabling large-scale patterns that span multiple world cells. Figure~\ref{fig:combined_world_gen} (right) shows this strongest form of control: a circular water pattern spanning four cells is used to generate an island-like structure in the center of the map, and a volcanic mountain pattern extends the map vertically.

% Together, these demonstrations show how a single block-level diffusion model supports a progression from unconditional chunk synthesis to precise user-guided world creation beyond the native $32^3$ training window. Notably, this capability requires no additional training, architectural changes, or computational overhead beyond running the same model repeatedly. While the sliding-window procedure does not incorporate true global context, the local spatial priors learned during training on $32^3$ chunks allow information to propagate across cells through their shared overlap regions, producing large-scale coherence from purely local generation.

\section{Human preference study}
\label{sec:user_study}
Downstream co-creative tools and procedural content generation pipelines only work if humans find the generator's output plausible and compelling. For interactive game environments like Minecraft, humans are ultimately the most important evaluators of generated content. We conduct a small-scale human preference study to study two questions: (i) Are our generated Minecraft chunks comparable in quality to real Minecraft terrain? (ii) Is render-based FID aligned with human preference? We recruited 19 participants with Minecraft experience to complete 2-alternative forced choice trials. In each trial, users are asked to compare two sets of biome-conditioned chunks drawn from different sources: MD4 (patch 2), MD4 (patch 4), DDPM (patch 2), and real validation chunks. Users select the set with higher quality and biome adherence, with free pan/zoom/rotate control over the visualization. Full instructions, interface, and data are provided in Appendix \ref{sec:apdx_user_study}.

\paragraph{Generated vs. real chunks} 

Human raters preferred generated chunks to real chunks across all three models (Table \ref{tab:user-study-pairwise-list}), with both MD4 models reaching statistical significance under a one-sided binomial test ($p < 0.001$ and $p = 0.042$. respectively). We interpret this as a consequence of classifier-free guidance, which steers outputs towards prototypical examples for the conditional biome. In contrast, real chunks are sampled from the Minecraft world, and may contain chunks that blend multiple biomes or have less interesting features. Asking raters to select the set that better matches the biome may reward our model's mode-seeking behavior, leading to the significant preference rates of generated data.

\begin{wraptable}{l}{0.48\textwidth}
\caption{Pairwise preference matchups. Each row reports the percent of trials where source A was preferred over source B.}
\vspace{2mm}
\label{tab:user-study-pairwise-list}
\centering
\begin{tabular}{lllr}
Source A & Source B & Win rate & \# trials \\
\midrule
MD4 p2 & Real & \textbf{67.1\%} & 173 \\
MD4 p4 & Real & 57.1\% & 163 \\
DDPM p2 & Real & 55.2\% & 192 \\
\midrule
MD4 p2 & DDPM p2 & 49.4\% & 158 \\
MD4 p2 & MD4 p4 & \textbf{55.2\%} & 174 \\
DDPM p2 & MD4 p4 & 53.3\% & 180 \\
\bottomrule
\end{tabular}

\caption{Human agreement with FID conditioned on absolute FID gap between models in model-vs-model trials. $p$-values are one-sided binomial tests against 0.5.}
\vspace{2mm}
\label{tab:fid-alignment}
\centering
\begin{tabular}{llll}
Min FID gap & Agreement & $n$ & $p$ \\
\midrule
0 & 54.3\% & 512 &  0.0286 \\
5 & 61.7\% & 227 &  $<0.001$ \\
10 & 62.9\% & 159 &  $<0.001$ \\
15 & 66.1\% & 109 &  $<0.001$ \\
\bottomrule
\end{tabular}
\end{wraptable}

\paragraph{Model vs. Model}
Our patch 2 models are statistically indistinguishable in head-to-head matchups, and both outperform the patch 4 model, suggesting that the additional resolution afforded by lower patch size has a larger impact on generation quality rather than the diffusion formulation. Per-biome preference rates (Appendix \ref{sec:apdx_user_study}) show notable differences, where the patch 4 model falls to a 16.1\% win rate in the village biome, aligning with the FID discrepancy observed in figure \ref{fig:md4_vs_md4_vs_ddpm}. 

\paragraph{FID alignment}
Our trial data uses the same generated samples used to compute per-biome FID scores, enabling analysis of preference rates under these FID values. We filter trials to only the 512 model-vs-model comparisons. Across these trials, humans selected the lower FID set $54.3\%$ of the time ($p=0.029$), showing a weak but measurable alignment. Complicating this analysis is the fact that the majority of trials have a $\leq 5$ point difference in per-biome FID; limiting analysis to only trials with larger FID gaps, agreement between humans and FID rises monotonically (Table \ref{tab:fid-alignment}. We view this as suggestive evidence that FID is more informative when gaps are large, but larger studies that cover a spectrum of FID differences are needed to confirm this claim.

\label{sec4:experiments}

\section{Related Work}
\paragraph{Discrete Diffusion}
Diffusion models were first extended to discrete state spaces by \citet{hoogeboom2021argmax}, who introduced multinomial diffusion for categorical data. \citeauthor{d3pm} further generalizes the DDPM framework to discrete data with D3PM, which includes absorbing corruption processes that replace discrete data with a \texttt{[MASK]} token. \citet{md4} (MD4) and \citet{mdlm} (MDLM) concurrently simplify and extend the masked diffusion paradigm, deriving simplified training objectives and demonstrating the effectiveness of masked diffusion for text and image generation. \citet{nie2025large} scale masked diffusion in the domain of language generation, scaling to 8 billion parameter models that match autoregressive baselines. Discrete diffusion is often applied to image generation, where images are first tokenized into discrete codes via a learned quantization model \cite{bond2022unleashing, esser2020taming}. In these settings, the continuous image data must first be transformed into these discrete learned representations. A key motivation for our work is that this intermediate tokenization step, which comes with the additional encoding and decoding overhead, can be bypassed entirely when working with natively discrete data. This property, which is used for controllable text infilling in SEDD, is central to the controllability results we demonstrate in the 3D voxel domain of Minecraft \cite{lou2023discrete}.

\paragraph{3D Diffusion} There have been many successful applications of diffusion models to 3D data generation, though most work targets continuous representations. Early approaches applied diffusion directly to point clouds and hybrid point-voxel representations \cite{luo2021diffusion, zhou20213d}. Recent methods, such as XCube, operate in learned latent spaces and support large effective spatial resolutions using sparse voxel representations \cite{ren2024xcube}. While the voxels output by these methods are structurally akin to Minecraft blocks, they do not carry the categorical, semantically meaningful labels like blocks do. Our work differs in this regard, operating under a finite vocabulary that encodes both appearance and gameplay function in each token.

\paragraph{Generative AI in Minecraft}
Minecraft has also served as a testbed for a variety of generative approaches. WorldGAN~\citep{worldgan} uses a 3D GAN trained on a single exemplar chunk, introducing \emph{block2vec} embeddings to handle the large block vocabulary. While this approach tackles a similar terrain-generation problem as our work, trained models learn to mimic the style of a single chunk rather than learning the general terrain distribution. Most closely related to this work are the diffusion-based methods of Scaffold Diffusion \cite{jung2025scaffold} and PERSIST \cite{garcin2026beyond}. Scaffold diffusion similarly treats voxels as tokens, operating directly in the block space to generate Minecraft structures, but conditions on an input occupancy ``scaffold'' that defines the structure geometry rather than generating from scratch. PERSIST, concurrent to this work, uses a 3D DiT-based rectified flow matching model to generate latent voxel representations of terrain in a Minecraft-like world. While this enables controllable editing of latent voxel worlds, it serves as a component of a video generation system rather than a standalone generator. Our work differs from these in scope, working at the native resolution using data collected directly from Minecraft.

\label{sec5:related}

\section{Limitations}
\paragraph{Evaluation}
Our primary metric, render-based FID, inherently struggles to assess internal structure (e.g., whether a building has a coherent interior), subsurface features (e.g., cave systems), or functional validity (e.g., whether doors are reachable). FID is typically computed using tens of thousands of generated samples. Computing FID on only 1,500 images per model introduces variance into our FID values, and the $\approx 60$ GPU-hours required for inference prevents its use as a train-time or model selection metric. The high inference cost inherent to diffusion sampling limits the scale of both our quantitative and qualitative evaluations, particularly for world-generation experiments, with each $5 \times 5$ world requiring over an hour of inference time on an H100 GPU.

\paragraph{Scope of the dataset} \implname is collected from Minecraft version 1.12.2, due to its compatibility with existing tools build upon Evocraft \cite{grbic2021evocraft}. This version was released in 2017, and subsequent updates have introduced new biomes, blocks, and structured content that our dataset does not capture. Our vocabulary compression within this dataset limits the expressivity of the data, which is particularly noticeable for human-authored data sources. We hope that further tools can build upon the data collection pipelines we create, and enable generative modeling at direct one-to-one correspondence with the native world representation. 

\paragraph{User study}
Several factors limit the strength of conclusions drawn from our human preference study. Our sample size is small, with 19 participants all recruited from within our institution, resulting in $\approx1,000$ total comparison trials. All users report familiarity with Minecraft, but we do not quantify or stratify by experience level, which may have a large impact on their evaluations. Our FID alignment analysis assumes that per-biome FID scores are representative of the four chunks shown in each trial, ignoring chunk-level variance. Our study design only evaluates biome-conditioned chunk generation, and we do not investigate the controllable inpainting and outpainting capabilities. While we find some suggestive evidence that FID aligns with our participants' preference when FID differences are high, agreement only reaches $66.1\%$ at maximum. Further studies on human evaluation in this domain and development of human-aligned metrics for quality remain an open problems.

\section{Broader Impacts}
Training generative models on human-authored data raises important ethical considerations. The human-authored maps we use were created by a professional map-builder, who earns income from their work. Generative models capable of recreating such content, if used irresponsibly, could undermine a rich community-based creative industry. While we obtained explicit permission from the creator to use and release this data for research purposes, we acknowledge that advancements in this field raise important questions about artistic integrity and attribution. 

\section{Conclusion}
We introduce \implname, a large-scale block-resolution dataset of Minecraft terrain, and use it to conduct the first study of controllable 3D diffusion models that operate directly in the space of Minecraft blocks. We find that both discrete and continuous diffusion models are capable of modeling this distribution, with generated output comparable in quality to real Minecraft terrain according to humans. While both approaches are competitive in this regard, we demonstrate how the discrete masked formulation yields naturally controllable generative models. Inpainting, outpainting, and user-authored block prompting all arise as emergent capabilities of discrete masked diffusion models trained with a simple class-conditional generation objective. We experiment extensively with these capabilities, illustrating their potential to empower human creators both by providing new modes of creative expression and enabling controllable interactive environment generation. Our experiments show that design choices at the dataset and architecture level have substantial impacts on generation quality. Smaller patch sizes in the underlying Transformer model prove better for modeling finer structural details, and generalize better to controllable generation modes. Targeted oversampling for complex data also improves generation performance, motivating the targeted scraping and curation pipelines used to create \implname. We release the full \implname dataset, pretrained models, and all training and data collection code to further the study of generative modeling in interactive, compositional 3D domains.

\section*{Acknowledgements}
We thank Timenth\_Buildings for generously providing access to his professionally crafted Minecraft maps and granting permission for their use in this research. We acknowledge NYU High Performance Computing for providing computational resources used to train and evaluate our models. This work was initiated during an internship at Sakana AI, and we gratefully acknowledge their support and compute resources

% \clearpage
\bibliographystyle{plainnat}
\bibliography{main}

\newpage
\appendix
\section{Appendix}

\subsection{Discussion}
\label{sec6:discussion_conclusion}
\paragraph{Blocks as tokens.}
A recurring theme across our results is that operating directly on Minecraft's native block representation unlocks capabilities that would be difficult to achieve in pixel or latent space. Exact inpainting, block-level conditioning from user-authored prompts, and guaranteed preservation of context during sliding-window outpainting all follow from the one-to-one correspondence between our diffusion representation and world voxels. This alignment, and the downstream controllable generation modes it unlocks are a core insight of our work.

A consequence of this is that our most compelling results are inherently qualitative. It is difficult to imagine a metric to compute the ``volcano-ness'' of a generated output, given a human-authored block context that suggests a volcano as seen in Figure \ref{fig:manual_context_infill}. We rely on visual demonstrations to convey these capabilities, and acknowledge the limitations in evaluating generative quality for these human-in-the-loop methods. Developing robust metrics for controllability in 3D generation remains an open problem.

A practical benefit of this is that our model can both read and write to Minecraft directly. Generated chunks are valid pieces of the world representation, and can be pasted into a live world via modding tools. Similarly, existing Minecraft worlds are already in the representation our models accept as conditioning. This enables in-game co-creative generation and editing, where players can control the generative process by simply placing blocks.

\paragraph{Discrete vs. Continuous Diffusion}
Our experiments show that both discrete and continuous diffusion paradigms are capable of producing high quality terrain chunks in this domain. Under identical conditions, continuous and discrete diffusion models achieve nearly identical performance, under both render-based FID and human preference. DDPM requires a finer spatial resolution to achieve this in our setup, and we only successfully train models at patch size 2. MD4 remains competitive at path sizes 2 and 4, with lower patch sizes providing significant improvements in training and inference time. The discrete masking formulation also enables simple implementation of the controllable generation methods. Inpainting with DDPM models is a common use-case in image generation, and many techniques \cite{controlnet, lugmayr2022repaint, song2020score} have addressed the schedule mismatch problem. We focus on the discrete case for the simplicity of implementation that MD4 affords, which uses the same sampling process as biome-conditioned generation. We investigate the analogous DDPM sampling process, injecting appropriately noised embeddings during the sampling process, in section \ref{apdx:ddpm_inpainting}. The comparison between formulations we present should not be considered an exhaustive comparison of discrete vs. continuous formulations in this domain, and we hope that the public release of our code and data will facilitate further research in this area. 

Continuous space also introduces a design decision absent from the discrete setting: how to embed block types. We use a text-based embedding, treating block name similarity as a semantic prior, but this is one of many possible choices. Co-occurrence-based embeddings, such as the \emph{block2vec} representations used in WorldGAN, could provide complementary structural priors~\citep{worldgan}. 

A common strategy in diffusion modeling is to operate in a learned latent space, which provides efficiency gains through spatial compression of the data. This has proven highly effective across multiple modalities, and has been successfully employed many popular image generation pipelines \cite{stablediffusion, dalle2, vqdiffusion}. We make the deliberate choice to forgo spatial compression, trading efficiency for the controllability described above. The discrete nature of Minecraft also allows us to skip the discrete quantization step required for discrete diffusion on continuous data, treating Minecraft blocks directly as tokens. 

\paragraph{Data curation} Our data composition experiments show that naive data collection (i.e sampling chunks at their natural frequency in Minecraft) is inadequate for learning to model structurally complex content. Biomes like villages and caves suffer the largest degradation in quality (as measured by FID) compared to our curated balanced dataset. This motivates dedicated collection pipelines that we develop and release as part of \implname, and suggest that scaling dataset size alone is insufficient for this domain. While sampling from a procedurally generated source is essentially free, more interesting content (like human-made builds) is in finite supply, making this subject particularly relevant for further efforts.

\subsection{Future Work}
\paragraph{Applications}
The methods we present here have several applications: co-creative design tools, controllable PCG pipelines for infinitely replayable games, and synthetic environment generators for game-playing and world modeling research. A practical barrier to all of these is inference speed. Amongst generative models, diffusion models come with a much larger inference overhead due to the iterative sampling procedure. Our patch 2 models take $\sim 2.5$ minutes to create a single $32^2$ chunk, and patch 4 models take $\sim 25$ seconds. World-scale generation scales linearly with the size of the map, as cells must be generated in sequence, introducing a bottleneck to large-scale generation. While offline applications are more tolerant of this cost, co-creative tools may require smaller models or using distillation techniques to become feasible. We view this as an important direction for future work.

% \section{Future Work}

\paragraph{Text-conditional generation}
A natural extension of this work is fully open-ended text-conditional generation of Minecraft worlds. Biome labels are a first step in controllability, and can be easily extracted from existing Minecraft worlds. However, the semantic gap between biome labels and rich natural language descriptions is significant. Closing this gap requires millions of individually captioned chunks that capture fine-grained details of terrain and structure, and could prove prohibitively expensive to create by hand.

A promising direction is to leverage the implicit Minecraft domain knowledge encoded in modern Large Vision-Language Models. Existing models are capable of visually parsing and reasoning about the game, as demonstrated in JARVIS-1 \cite{wang2023jarvis1}, and this knowledge could be applied to automatically generate synthetic captions, similar to the approach employed in Moonshine \cite{nie2025moonshine}. A limitation of this is approach is that current frontier models accept only image inputs. Render-based captioning is thus limited to only capturing details the model can ``see'', ignoring features underneath. Until language models accept 3D voxel data directly as an input modality, synthetic labeling may require more nuanced strategies than single-view image captioning.

\subsection{Human-authored data}
\label{apdx:human_Data}
\begin{wraptable}[14]{r}{0.48\textwidth}
\caption{Chunk counts for each human-authored map}
\vspace{4mm}
\label{tab:human_chunk_count}
\centering
\begin{tabular}{lr}
Map & Count \\
\midrule
Ancient Empire & 87,917 \\
Kingdom of Sarano & 36,768 \\
Marethea & 33,050 \\
Osirion & 103,619 \\
Sora Kingdom & 93,030 \\
The 5 Bridges & 58,007 \\
\bottomrule
\end{tabular}
\end{wraptable}
\implname contains chunks from six professionally made Minecraft maps, which contain much more complex structured content than procedurally generated content. Table \ref{tab:human_chunk_count} shows the names of each map, and the number of chunks collected. We train MD4 and DDPM (patch 2) models on a balanced dataset, including all 15 natural biomes plus the six human map biomes. We bootstrap existing chunks for the Kingdom of Sarano, Marethea, and The 5 Bridges maps which lack sufficient chunks for fully balanced representation. Table \ref{tab:natural_v_human_augmented} shows the per-biome FID scores for the natural biome and human-augmented models. Natural biome models typically outperform on the original set of 15 biomes, potentially due the higher number of samples in these biome classes ($\sim$ 67k vs. $\sim$ 47k). The DDPM model trained on human data achieves significantly lower FID in five of the 6 maps, though we note the human-authored data introduces significantly higher variance into this metric, as human-authored chunks represent more diverse content than natural biome chunks both in vocabulary and structure. 
\begin{table*}[ht]
\centering
\caption{Per-biome FID comparison between MD4 models trained with human data vs. without. Lower is better. Bold indicates the lowest available FID for each biome. A dash indicates the model has no value for that biome.}
\label{tab:natural_v_human_augmented}
\small
\begin{tabular}{lcccc}
\toprule
Biome & MD4 p2 natural & MD4 p2 +human & DDPM p2 natural & DDPM p2 +human \\
\midrule
Beaches & 63.49 & \textbf{60.97} & 74.21 & 85.87 \\
Birch forest & \textbf{41.83} & 48.59 & 45.28 & 49.27 \\
Cave & \textbf{46.52} & 46.69 & 55.98 & 51.96 \\
Desert & 57.94 & 58.51 & \textbf{52.42} & 60.90 \\
Extreme hills & 103.78 & 97.65 & \textbf{86.13} & 87.30 \\
Forest & 72.84 & 54.30 & \textbf{49.39} & 53.56 \\
Ice & \textbf{62.11} & 63.98 & 72.30 & 66.73 \\
Jungle & \textbf{60.21} & 66.99 & 61.42 & 79.56 \\
Ocean & 55.17 & 59.13 & 34.66 & \textbf{27.40} \\
Plains & \textbf{39.31} & 74.59 & 46.02 & 77.69 \\
River & \textbf{54.81} & 64.50 & 59.48 & 62.10 \\
Savanna & 57.60 & 89.17 & \textbf{54.32} & 96.66 \\
Swampland & \textbf{56.92} & 57.73 & 81.28 & 61.44 \\
Taiga & 54.71 & \textbf{52.30} & 57.80 & 62.08 \\
Village & 61.65 & 68.76 & \textbf{58.61} & 75.65 \\
Ancient empire & \textemdash & 149.38 & \textemdash & \textbf{118.42} \\
Kingdom of sarano & \textemdash & 156.53 & \textemdash & \textbf{113.45} \\
Marethea & \textemdash & \textbf{161.41} & \textemdash & 168.46 \\
Osirion & \textemdash & 114.01 & \textemdash & \textbf{106.34} \\
Sora kingdom & \textemdash & 177.43 & \textemdash & \textbf{139.41} \\
The 5 bridges & \textemdash & 118.90 & \textemdash & \textbf{112.88} \\
\midrule
\textbf{Average} & \textbf{59.26} & 87.69 & 59.29 & 83.67 \\
\bottomrule
\end{tabular}
\end{table*}

\begin{figure}[h]
    \centering
    \includegraphics[width=0.9\linewidth]{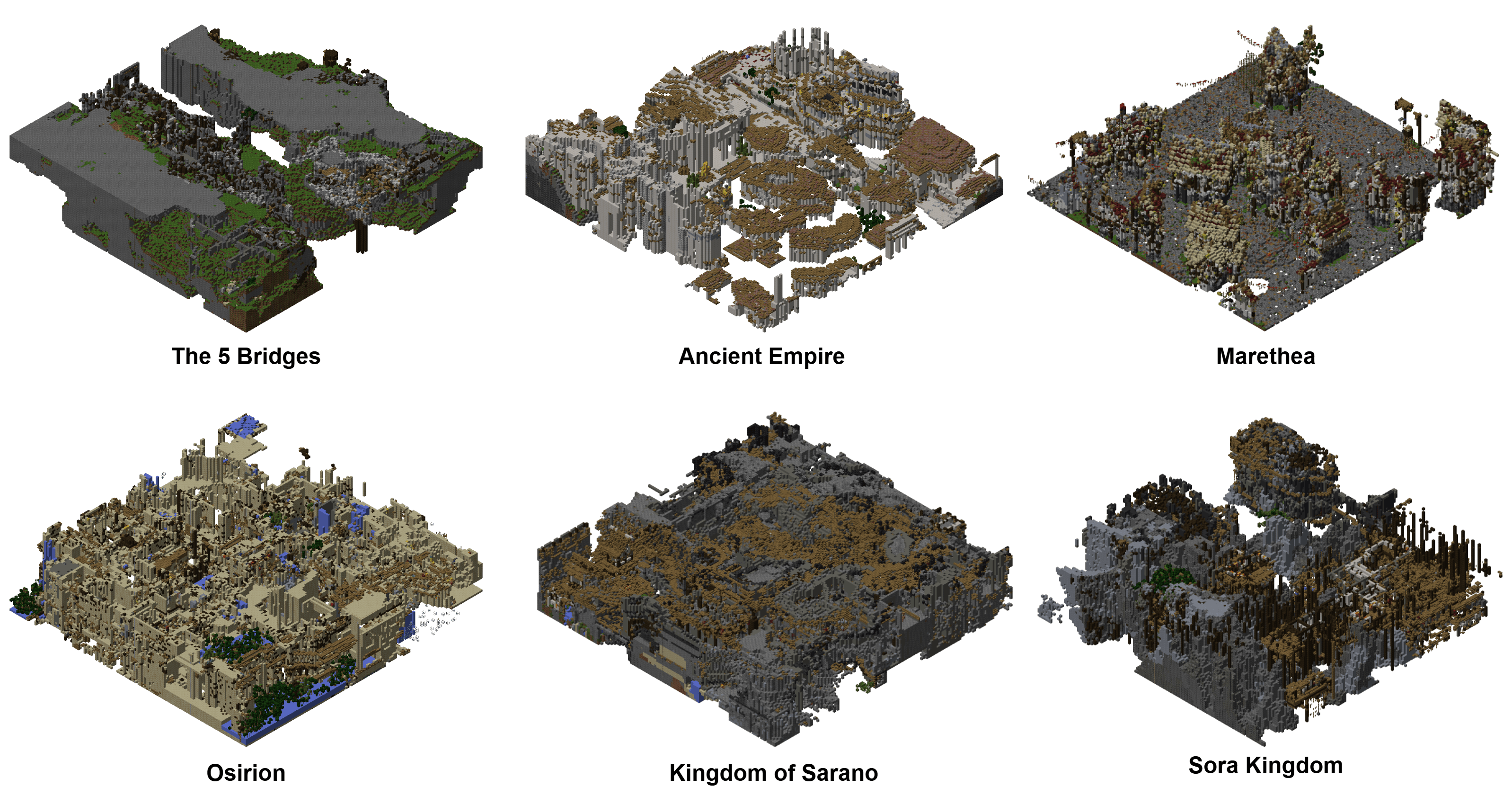}
    \caption{$5\times5\times1$ worlds generated by the MD4 model trained on the human-augmented dataset. Each world uses one of the six human-authored maps as biome condition for all cells}
    \label{fig:human_worlds}
\end{figure}

While our qualitative results show our models learning to generate recognizable architectural and stylistic features, they do not compare to the quality of the original map. This is in part due to our aggressive block compression scheme, which cuts down on token sparsity and vocabulary size. Decorative blocks such as buttons and trapdoors are painstakingly placed by map creators to add sub-voxel details to the map, and are entirely removed from our dataset. Collapsing to base block types also introduces a visual artefact---in the original map, a colored wool variant might be used to provide texture by blending into similarly colored blocks, while collapsing this to a base white wool texture creates jarring contrast and hides the original aesthetic style.  We view the generation of this complex structured data as an open challenge that our dataset enables. Accurately modeling human creation requires preservation of this sparse block vocabulary and overcoming the problem of metadata-induced vocabulary explosion.

Figure \ref{fig:human_worlds} shows biome-conditioned world generation using a model trained on a desert including the six human-authored chunks. When generating at the world scale, we see evidence of our model learning long-range coherence for these maps. The Ancient Empire and Marethea maps both show organized and separated buildings, with gaps suggesting roads between them. While we do not achieve the level of fidelity present in our natural biome dataset for these classes, this demonstrates the inherent challenge and trade-offs introduced when the data spans a large part of the true Minecraft block vocabulary. We discuss the implications and limitations of human-authored generation in Section \ref{sec6:discussion_conclusion}.

\subsection{Vocabulary processing}
\label{apdx:vocab}

\paragraph{Natural Biomes vocabulary} Our core dataset includes all naturally-occurring blocks present in the scraped data. The finally vocabulary used for training our models expands on this set by incorporating additional block states for select block types. Minecraft blocks can have associated state metadata that can affect their appearance and behavior in the game world, such as their relative orientation. While this metadata typically encodes minor aesthetic details, we explicitly handle metadata for ``slab'' and ``stair'' blocks. These block types are commonly used in Minecraft structures; Both occupy less than a full voxel and allow for sub-voxel details in structures. Rather than including an additional metadata channel, we fold all possible ``slab'' and ``stair'' states into our block vocabulary, remapping each unique orientation state to a unique block id (2 states for slabs, 48 states for stairs). To avoid an explosion in vocabulary size, we merge all possible materials for ``slabs'' and ``stairs'' into a single material type.

\paragraph{Human-authored vocabulary} User-created maps use a much broader palette than naturally-spawning Minecraft blocks. Naively incorporating all blocks and variations would make the number of unique block types increase by almost an order of magnitude to 1,181. Thus, to keep our vocabulary size manageable, we manually remove and remap certain block types to a compressed set. We remove rare decorative blocks, like ``buttons'' and ``rails'', by remapping them to the ``air'' block. For block types with many variants, like ``light\_blue\_wool`` and ``green\_wool'', we also remap all variants to a single base case. We manually identify these axes on which to compress our vocabulary, and use Claude Opus 4.6 to parse the full list of block names, identify block name substrings indicating variants, and automatically generate the mappings to the subset of base block types. After processing, the vocabulary across natural and human-authored data contains $V = 177$ unique block types.

\subsection{Data collection and curation}
\label{apdx:data_collection}

\subsubsection{Evocraft}
We collect chunks of procedurally generated Minecraft worlds by distributing data collection across a cluster. Each node runs an independent local server with Minecraft version 1.12.2 and extracts terrain chunks via a custom Evocraft-based API \cite{grbic2021evocraft}. We reset each node's world seed every 10,000 chunks, which serves to increase diversity and lower memory overheads. Evocraft provides a base python API for interfacing with a running Minecraft server, exposing methods for reading in volumes of blocks. We use this API to collect natural biome chunks, cave chunks, and village chunks. 

\paragraph{Natural biome collection} We traverse procedurally generated worlds in a square spiral pattern in the ground-level X-Z plane, starting at the world spawn. We stride our collection window in increments of 32 blocks, so that each collected sample corresponds to a unique chunk of size $32 \times 32 \times 32$. At each coordinate in the X-Z plane, we estimate the surface height by querying the highest Y coordinate of a non-air block. We center our 32-block window around this Y coordinate at that location, then add vertical diversity by shifting our window by $\pm 12$ blocks. While Minecraft defines 76 biomes in version 1.12.2, we note that many of these are variants of a base biome, e.g., ``jungle\_hills'' and ``jungle\_edge''. Following the same compression philosophy as our block vocabulary, we compress these variants into a single parent biome label, e.g., ``jungle'' (see \ref{tab:biome_mapping} for all remappings).  This data collection level accounts for a large bulk of \implname, yielding 1,521,781 labeled chunks across 13 parent biomes.

Due to the relative rarity of certain biomes, we additionally design a biome-targeted collection pipeline. Using PyuBiomes, we first identify $32^3$ areas of the map that contain a desired target biome, without requiring a full sweep over the terrain. We pass these coordinates to our Evocraft collection pipeline, which collects chunks from only these locations, enabling more efficient collection of rare biomes.

\paragraph{Underground collection}
To create the ``cave'' biome present in \implname, we modify our world traversal method to collect chunks from under the world surface. We use a similar square-spiral traversal as above: at each X-Z location traversed by our square-spiral extraction path, we scan downward in 32-block strides and identify caves by thresholding based on the presence of air voxels ($> 10\%$ of the volume). This data collection level yields a total of 146,000 cave chunks.

\paragraph{NPC Village collection}
To efficiently collect chunks from procedurally generated villages, we use the Pyubiomes \cite{pyubiomes} package, which exposes methods for computing the coordinates of villages based on a given world seed and Minecraft version. For each random seed, we find the coordinates of NPC villages within 10,000 blocks of spawn, and pass these coordinates to the Evocraft component. Around each village coordinate, we extract a large volume of blocks ($256 \times 64 \times 256$) from the world, and then densely extract chunks by striding a $32^3$ collection window with stride $4$. We filter these chunks using three criteria to ensure clean samples:
\begin{enumerate}
    \item We discard samples that are primarily underground via an air-content threshold (retaining chunks $> 15\%$ air).
    \item We discard chunks that contain less than 60 village-indicator blocks, such as doors and stairs, ensuring chunks contain structures, rather than natural terrain surrounding a village
    \item We discard samples where village indicator blocks touch the boundary of the volume, ensuring that our model learns to generate complete structures rather than truncated ones.
\end{enumerate}
Overall, this data collection level yields 178,271 village chunks. 

\paragraph{Human data collection}
We extract chunk data from human-authored by extending the ``mcaselector'' tool \citep{mcaselector}, to allow for targeted selection of regions specified from a top-down map view of the map. This ensures our scraping process is constrained to chunks containing the human-authored structures of interest, rather than the natural terrain from the base world seed. Unlike the natural overworld, these maps contain structures that extend hundreds of blocks upwards. To handle the increased verticality of these maps, we first identify all $32\times 32$-block columns that overlap with the human chunks. We then scan vertically with a stride of eight and extract all $32^3$-block chunks until we reach a chunk with $< 10\%$ air blocks. In total,  this data collection level yields 358,762 human-authored chunks.

\subsection{Biome label compression}
Table \ref{tab:biome_mapping} lists the source and target biome labels used in our biome compression scheme. ``Mesa'' and ``Mushroom'' biomes occur very rarely during data collection ($< 1\%$), and are not included in our dataset or compression scheme due to lack of samples.
% Biome label compression: 
\begin{table}[h]
\centering
\small
\caption{Collapsed mapping from Minecraft-native labels to final biome labels used in  \implname.}

\begin{tabular}{>{\raggedright\arraybackslash}p{0.6\linewidth} | c}
\toprule
\textbf{Source biome names} & \textbf{Target biome label} \\
\midrule
ocean, deep\_ocean, deep\_warm\_ocean & Ocean \\
\hline
desert, desert\_hills, mutated\_desert & Desert \\
\hline
beaches, stone\_beach & Beaches \\
\hline
cave & cave \\
\hline
extreme\_hills, extreme\_hills\_with\_trees, mutated\_extreme\_hills\_with\_trees, mutated\_extreme\_hills & Extreme hills \\
\hline
forest, forest\_hills, mutated\_forest, mutated\_roofed\_forest, roofed\_forest & Forest \\
\hline
birch\_forest, birch\_forest\_hills, mutated\_birch\_forest, mutated\_birch\_forest\_hills & Birch forest \\
\hline
plains, mutated\_plains & Plains \\
\hline
river & River \\
\hline
savanna, savanna\_rock, mutated\_savanna\_rock, mutated\_savanna  & Savanna \\
\hline
swampland, mutated\_swampland & Swampland \\
\hline
taiga, taiga\_hills, mutated\_redwood\_taiga, mutated\_redwood\_taiga\_hills, redwood\_taiga\_hills, mutated\_taiga, redwood\_taiga & Taiga \\
\hline
ice\_flats, ice\_mountain, cold\_beach, frozen\_river, mutated\_ice\_flats, mutated\_taiga\_cold, taiga\_cold, taiga\_cold\_hills, ice\_mountains & Ice \\
\hline
jungle, jungle\_edge, jungle\_hills, mutated\_jungle & Jungle \\
\bottomrule
\end{tabular}
\label{tab:biome_mapping}
\end{table}

\subsection{Natural Occurrence Dataset}
\label{apdx:natural_occurrence}
To create our natural occurrence dataset, we first compute an estimated distribution of biomes in randomly generated Minecraft worlds. 

\paragraph{Natural Biome occurrence} To compute the distribution of natural biome values (i.e, all biomes in Table \ref{tab:biome_mapping} besides \emph{village} and \emph{cave}), we count the number of times we encounter these biomes in procedurally generated worlds, using the default world generation settings in Minecraft v1.12.2. Using our data scraping pipeline, we stride across the world on the X-Z plane, with $s=16$, sampling the biome label at each block location. We continue up to 10,000 blocks from the world spawn, then regenerate the world using a random seed. We repeat this process 5 times, resulting in a total of $1,309,790$ samples. This provides an estimated percentage of the game world that our 13 native minecraft biomes occupy, and we construct a dataset of 1,000,000 chunks stratified according to this distribution before moving onto the village and cave processing steps.

\paragraph{Village natural occurrence} Our natural biome collection pipeline does not discard any naturally occurring villages it encounters, thus there are a number of chunks with a natural biome label that contain village structures. These chunks represent villages at their naturally spawning rate, removing the need for estimation. We filter . We apply the village-indicator block criteria to our naturally occurring chunks from the previous step, re-labling any chunk with $\geq 60$ indicator blocks as village. We do not apply the air threshold or volume-boundary criteria to these samples, as this dataset aims to recreate the naive approach of striding across generated worlds without dataset curation.

\paragraph{Cave natural occurrence} Each $32^2$ patch of the world yields a single natural biome chunk at the world \textit{surface}, but may contain many cave chunks underneath. Because the surface level of the world is variable, there is no clear way to estimate what the spawn rate of underground caves is for a given patch of the Minecraft world. Instead, we set the number of cave biomes to be equal to the number of relabeled village chunks, treating it as a similar custom biome.

Table \ref{tab:natural_biome_dist} shows the estimated natural biome distribution, the distribution after village and cave processing, and the final counts of our natural occurrence dataset after adjusting for the two additional classes.

\begin{table}[!ht]
\centering
\begin{tabular}{lccc}
\hline
Biome & World Percent & Final Percent & Final Chunk Count \\
\hline
beaches & 3.90\% & 3.84\% & 38,450 \\
birch\_forest & 2.81\% & 2.77\% & 27,704 \\
cave & -- & 0.71\% & 7,100 \\
desert & 5.89\% & 5.81\% & 58,069 \\
extreme\_hills & 7.18\% & 7.08\% & 70,787 \\
forest & 11.76\% & 11.59\% & 115,942 \\
ice & 7.39\% & 7.29\% & 72,858 \\
jungle & 1.83\% & 1.80\% & 18,042 \\
ocean & 35.65\% & 35.15\% & 351,473 \\
plains & 7.64\% & 7.53\% & 75,323 \\
river & 4.02\% & 3.96\% & 39,633 \\
savanna & 3.59\% & 3.53\% & 35,394 \\
swampland & 2.81\% & 2.77\% & 27,704 \\
taiga & 5.52\% & 5.44\% & 54,421 \\
villages & -- & 0.71\% & 7,100 \\
\hline
\end{tabular}
\caption{Estimated biome percentages and final biome distribution for the natural occurrence dataset.}
\label{tab:natural_biome_dist}
\end{table}

\clearpage
\subsection{Additional qualitative samples}
\begin{figure}[!ht]
    \centering
    \includegraphics[width=0.85\linewidth]{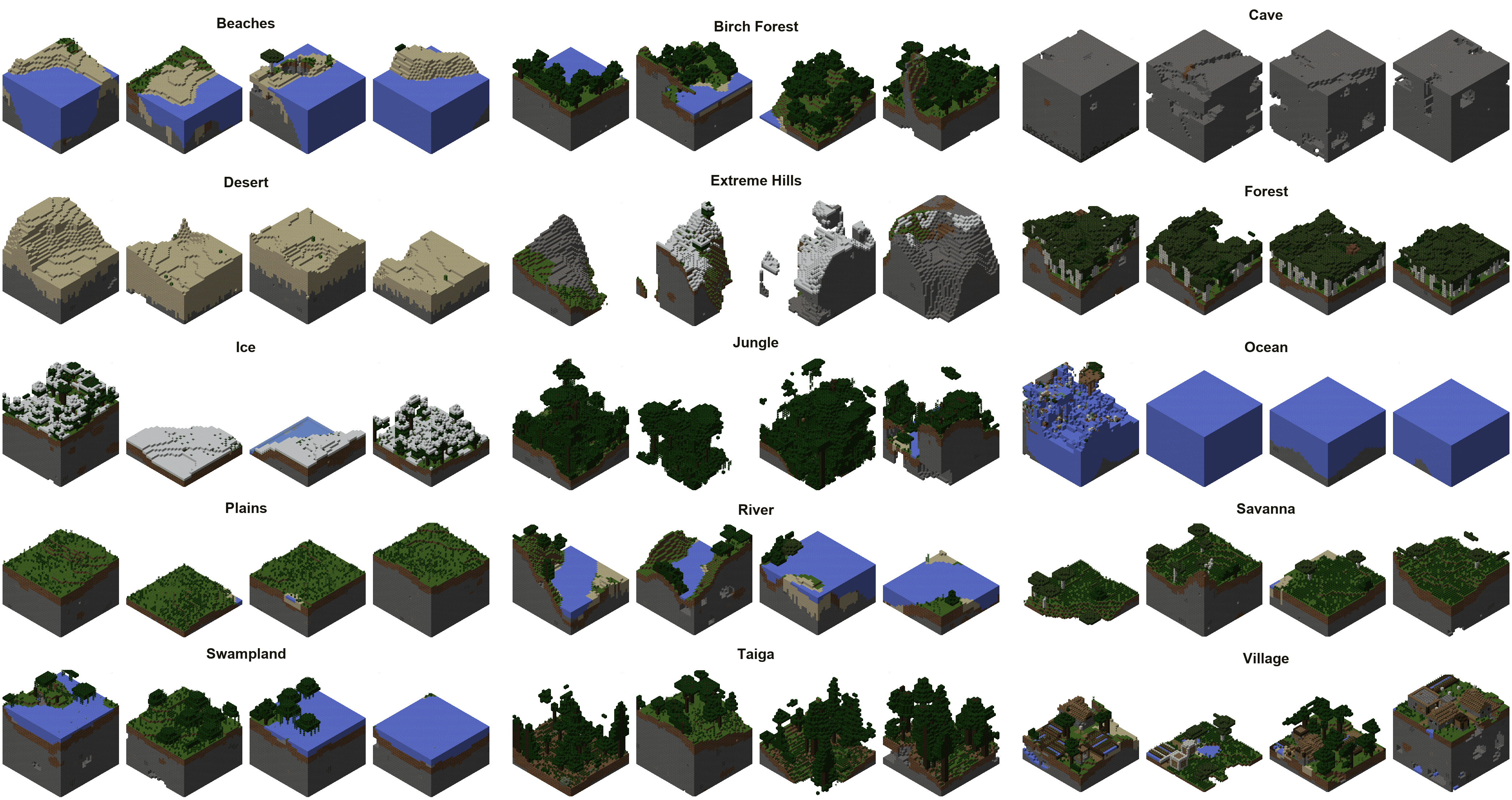}
    \caption{Randomly selected biome-conditioned samples generated by our MD4 (patch 2) model across all 15 natural biome classes.}
    \label{fig:MD4_biome_samples_full}
    \centering
    \includegraphics[width=0.85\linewidth]{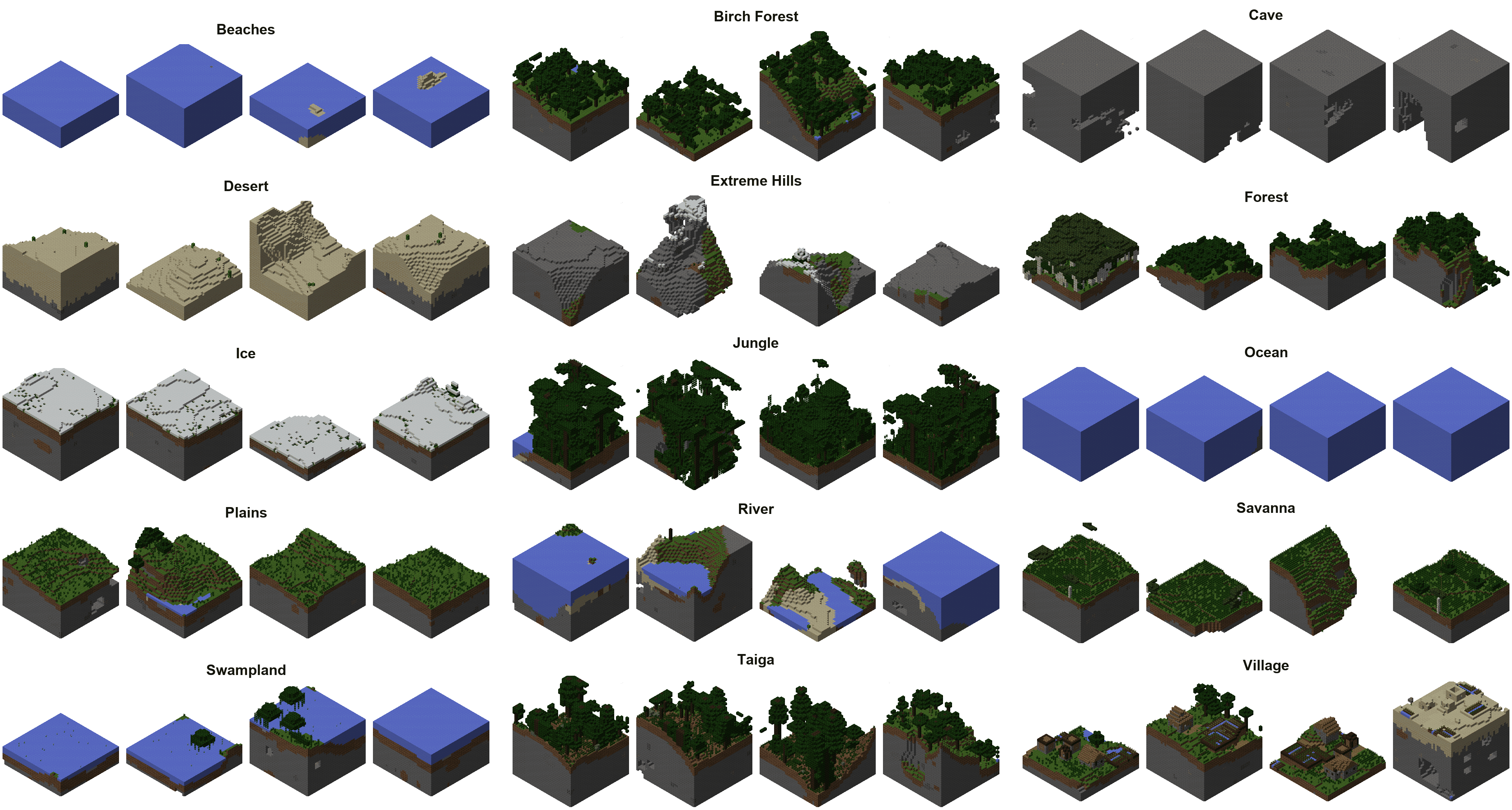}
    \caption{Randomly selected biome-conditioned samples generated by our DDPM (patch 2) model across all 15 natural biome classes.}
    \label{fig:DDPM_biome_samples_full}
\end{figure}

\begin{figure}[H]
    \centering
    \includegraphics[width=0.85\linewidth]{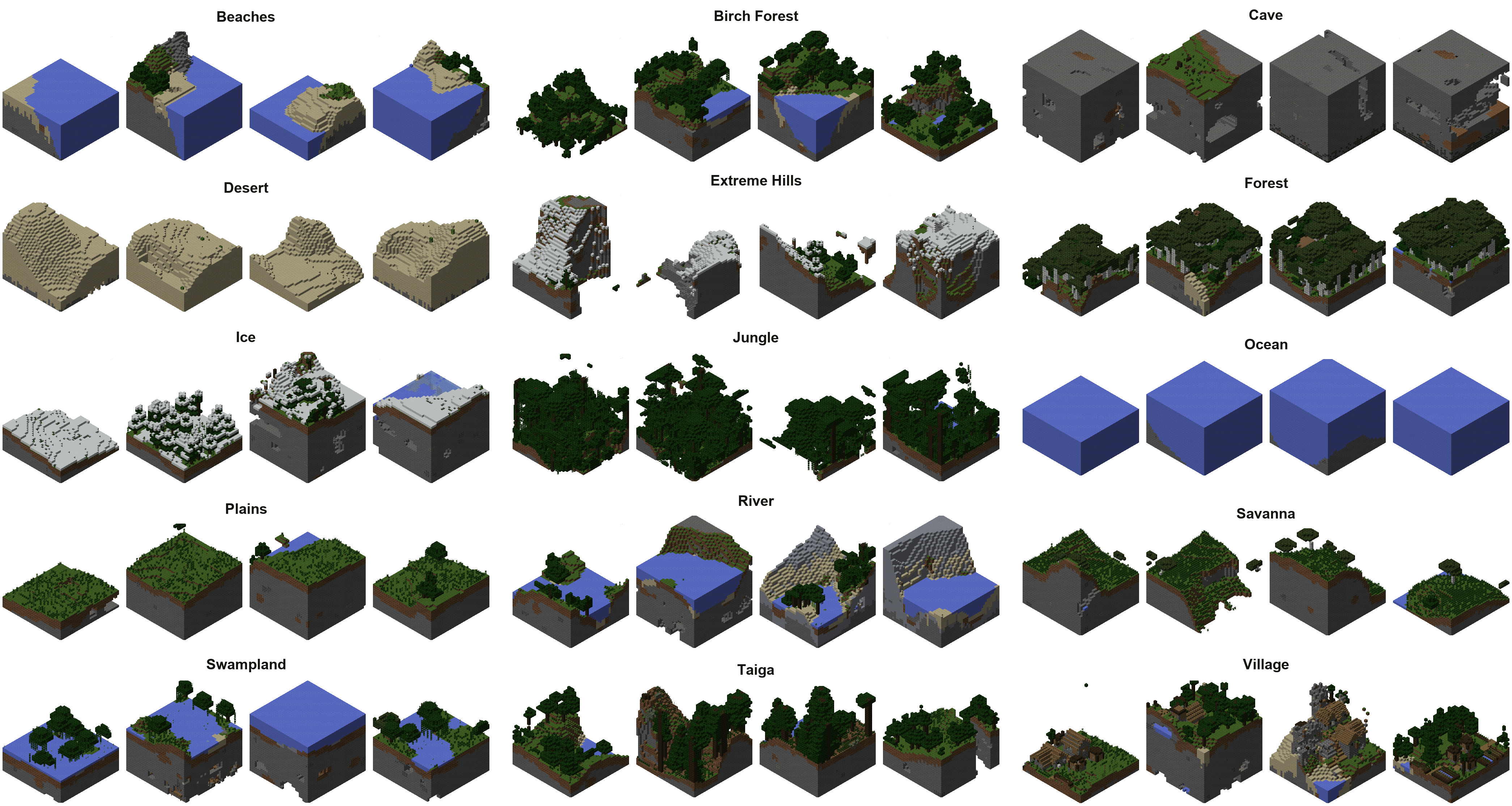}
    \caption{Randomly selected biome-conditioned samples generated by our MD4 (patch 4) model, trained on the \textbf{balanced} dataset, across all 15 natural biome classes}
    \label{fig:MD4_p4_balanced_samples}
    \centering
    \includegraphics[width=0.85\linewidth]{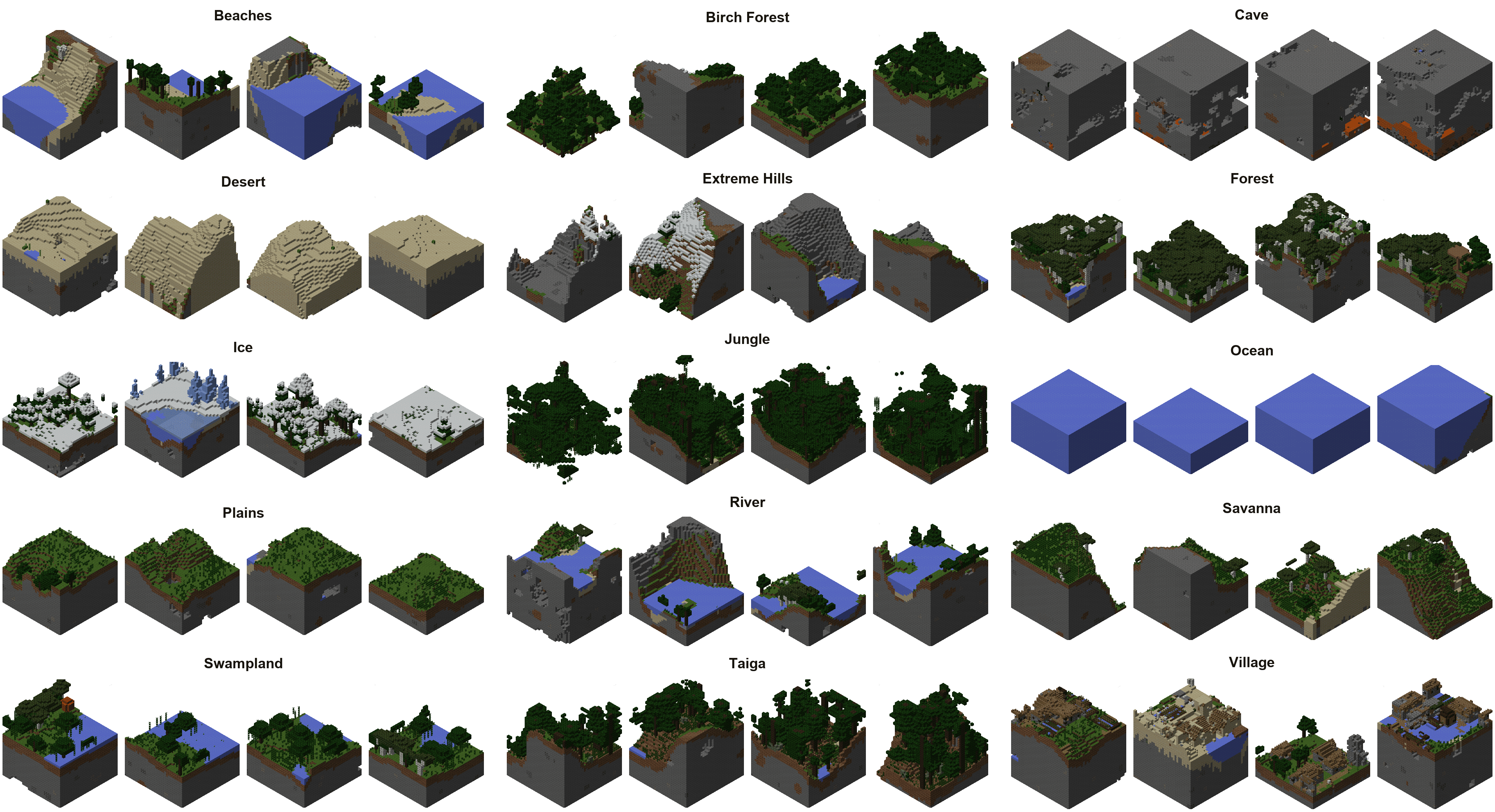}
    \caption{Randomly selected biome-conditioned samples generated by our MD4 (patch 4) model, trained on the \textbf{natural occurrence} dataset, across all 15 natural biome classes}
    \label{fig:MD4_p4_natural_samples}
    \centering
    \includegraphics[width=0.85\linewidth]{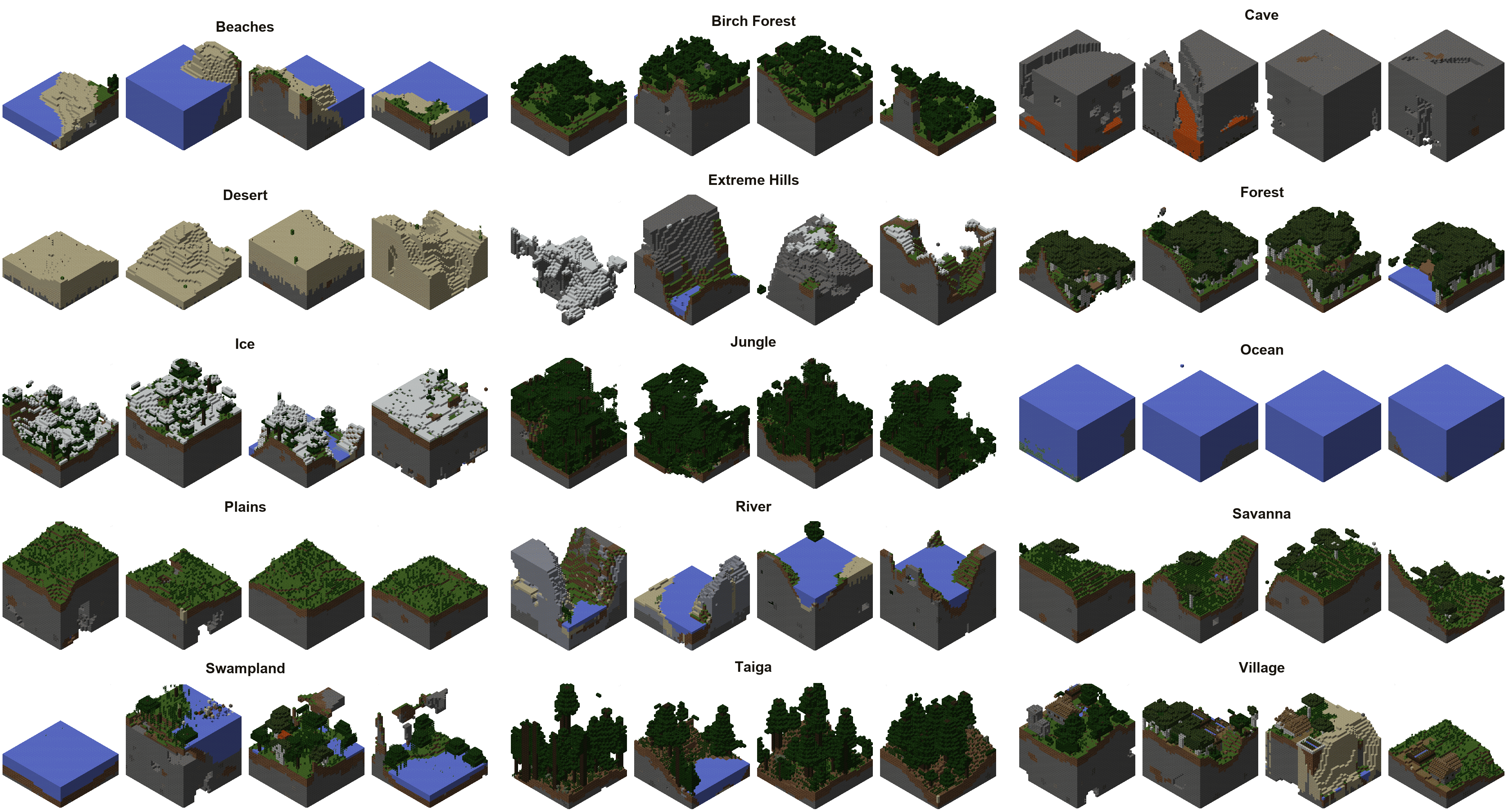}
    \caption{Randomly selected biome-conditioned samples generated by our MD4 (patch 4) model, trained on the \textbf{village boosted} dataset, across all 15 natural biome classes}
    \label{fig:MD4_p4_boosted_samples}
\end{figure}

\begin{figure*}[!ht]
    \centering
    \includegraphics[width=\linewidth]{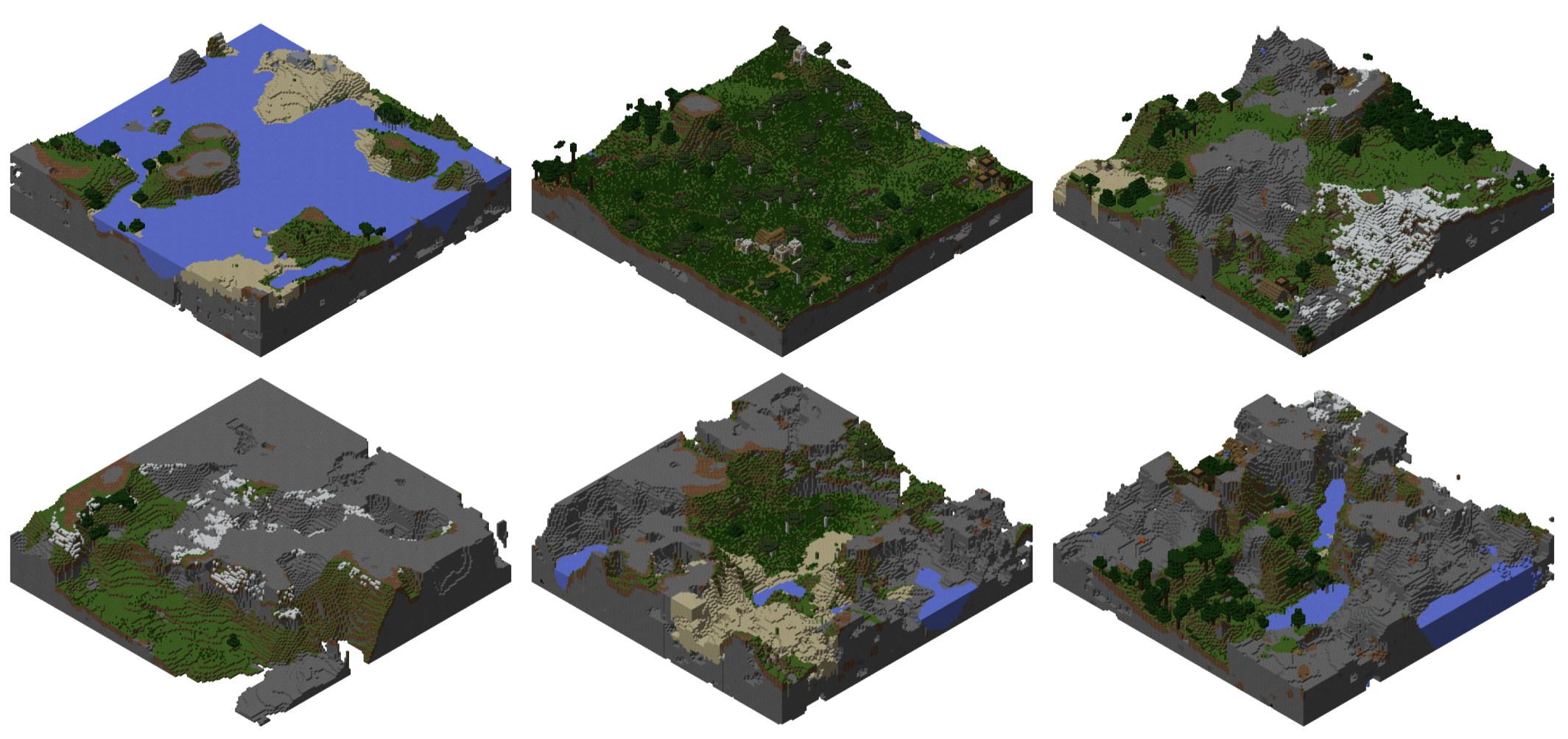}
    \caption{Six unconditional $5\times5\times1$ worlds generated by our MD4 (patch 2) model}
    \label{fig:Unconditional_worlds}
    \centering
    \includegraphics[width=\linewidth]{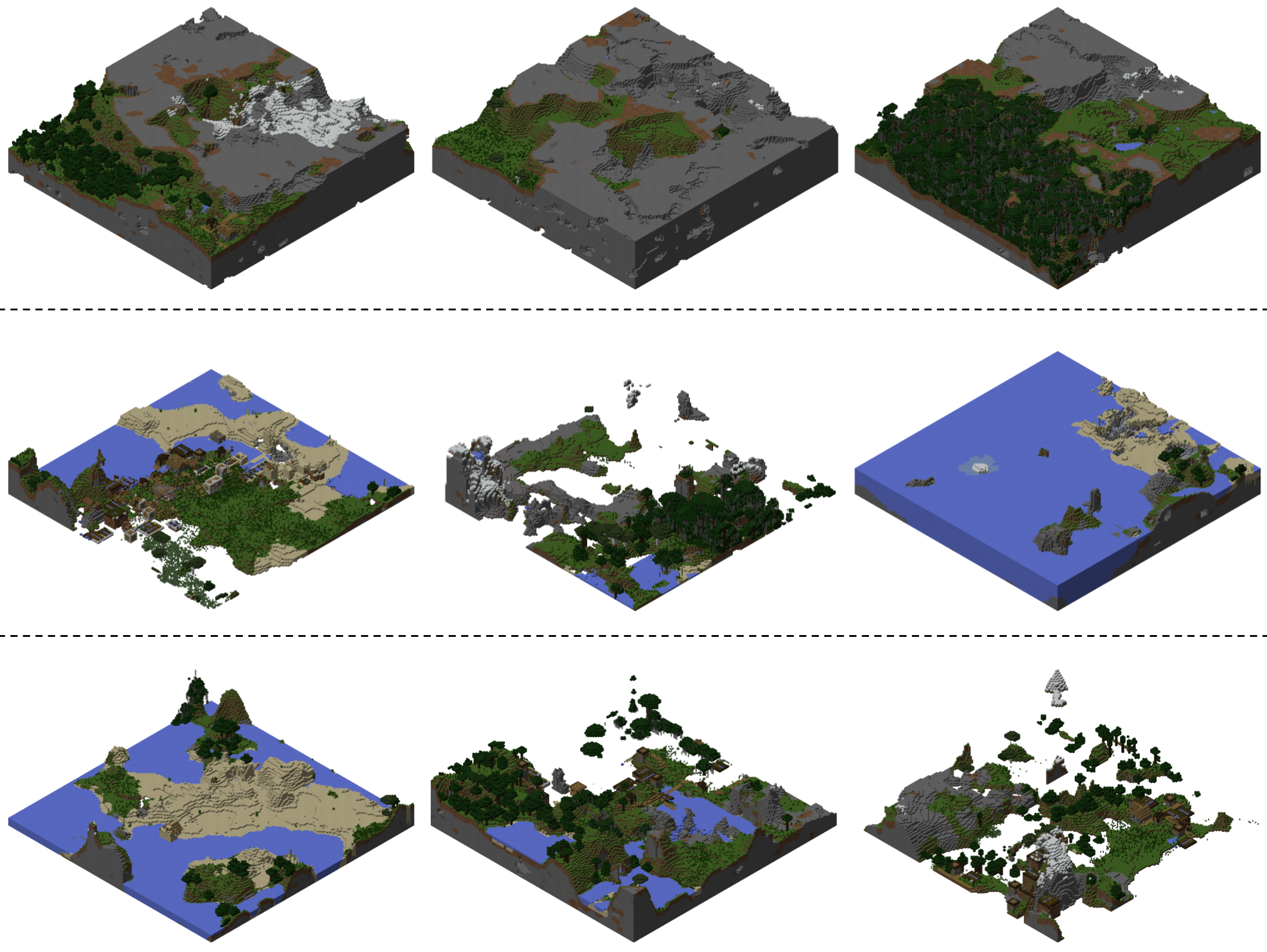}
    \caption{Three $5\times5\times1$ worlds generated by the three dataset variant MD4 (patch 4) models. Top row: balanced dataset. Middle: Natural occurrence dataset. Bottom: Village boosted dataset.}
    \label{fig:Unconditional_worlds_p4}
\end{figure*}

\begin{figure}[h!]
    \centering
    \includegraphics[width=\linewidth]{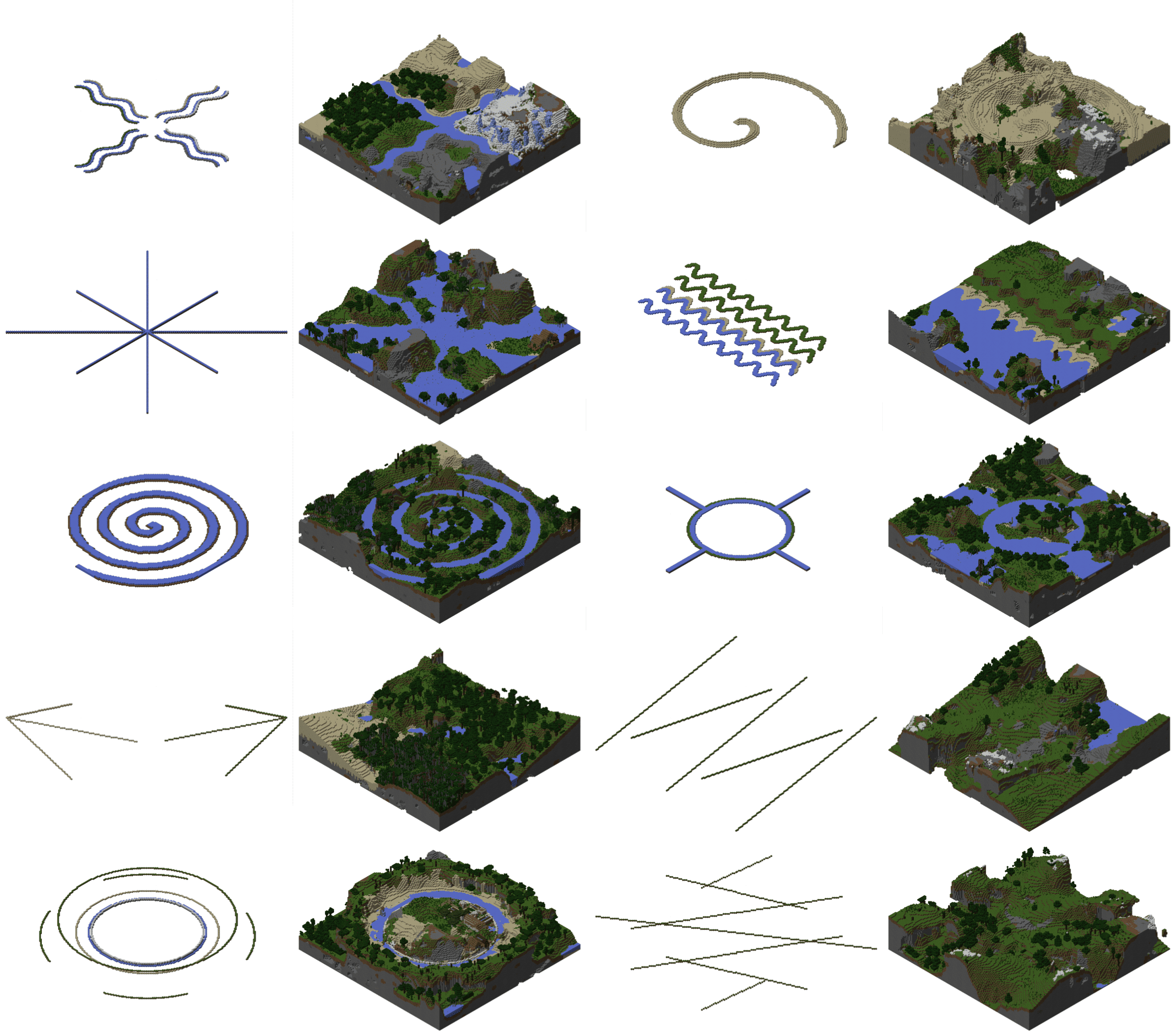}
    \caption{Block-conditioned $5\times5\times1$ worlds generated by our MD4 (patch 2) model. Each pair shows the context block pattern (left) and the corresponding generated world (right). Block patterns are automatically sliced to $32^3$ chunks and used as block context for inpainting, in addition to overlapping blocks from neighboring cells.}
    \label{fig:authored_worlds}
\end{figure}

Figures \ref{fig:MD4_biome_samples_full}, \ref{fig:DDPM_biome_samples_full}, \ref{fig:MD4_p4_balanced_samples}, \ref{fig:MD4_p4_natural_samples}, and \ref{fig:MD4_p4_boosted_samples} show biome-conditioned generated samples for the five models we present in this paper. Figure \ref{fig:Unconditional_worlds} shows six additional unconditional world generation samples using our MD4 patch 2 model. Figure \ref{fig:Unconditional_worlds_p4} shows three unconditional worlds generated by each dataset variants MD4 patch 4 model. Figure \ref{fig:authored_worlds} shows block contexts and resulting worlds in the user-guided world generation setting.

\clearpage
\clearpage
\subsection{Controllable Generation at larger patch sizes}
\begin{figure}[h!]
    \centering
    \includegraphics[width=\textwidth]{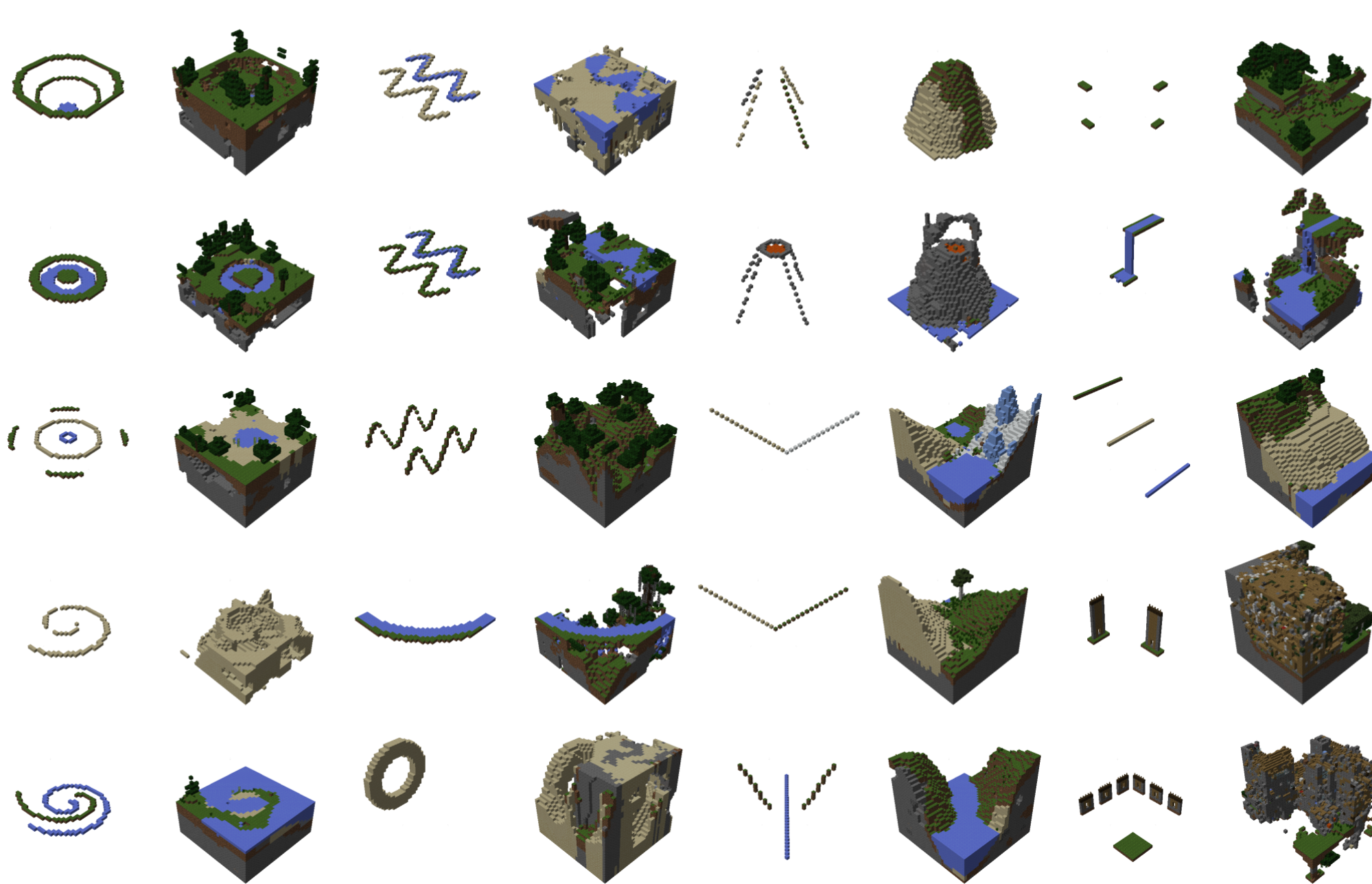}
    \caption{Block-conditioned inpainting using the MD4 patch 4 model. We generate 4 variants for each context pattern, and show the best looking generation for each. In cases where all variants were of high quality, we select the one most resembling the patch 2 version shown in figure \ref{fig:manual_context_infill}}
    \label{fig:p4_manual_context_infill}
\end{figure}

We observe degraded performance in block-conditional infilling using our patch 4 models. Figure \ref{fig:p4_manual_context_infill} shows the output of our user-authored block patterns using the balanced dataset MD4 patch 4 model. Some patterns result in generations of similar quality to the patch 2 model (shown in figure \ref{fig:manual_context_infill}), while others fail to generate coherent terrain. In particular, the contexts using village-specific blocks (rightmost, bottom two rows) generate outputs that bear no resemblance to the Minecraft terrain.

\subsection{DDPM Inpainting}
\label{apdx:ddpm_inpainting}
We evaluate our DDPM (patch 2) model in the standard inpainting and block-conditioned inpainting scenarios. We implement the replacement method introduced in \citet{song2020score}, injecting partially noised block embeddings for the fixed block tokens at every sampling step. We differ from their implementation, selecting starting timestep based on the number of blocks present in the context, rather than starting from $T$, matching our MD4 inpainting procedure. We add a final replacement step at the end, ensuring the context blocks are not decoded to different discrete block IDs.

\begin{figure}[h]
    \centering
    \includegraphics[width=0.9\linewidth]{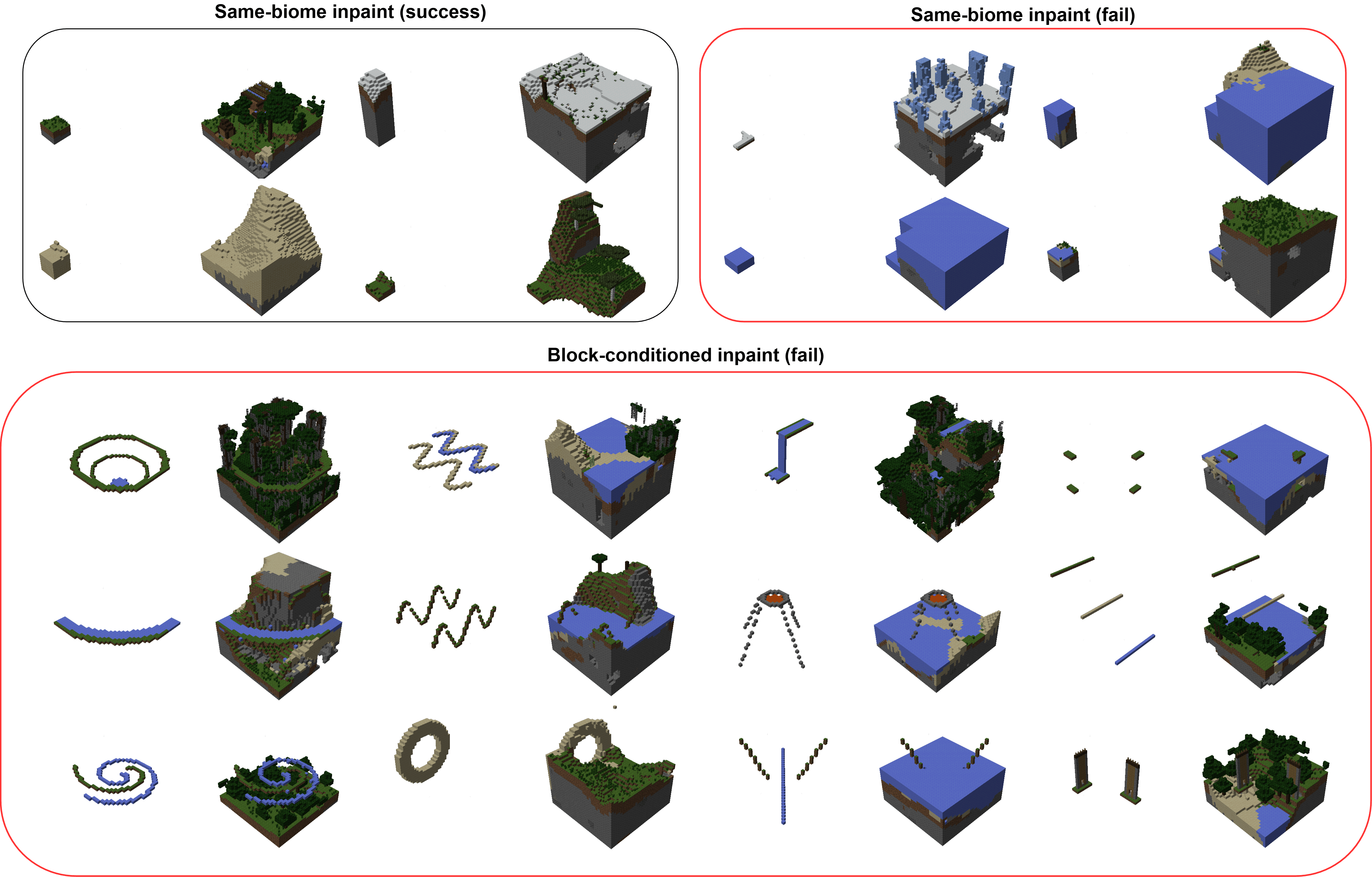}
    \caption{Standard inpainting from a $8\times8\times32$ chunk and from human-authored block contexts using our DDPM (patch 2) model.}
    \label{fig:ddpm_inpaint}
\end{figure}

Figure \ref{fig:ddpm_inpaint} shows the result of this experiment. In the standard inpainting scenario, our DDPM model is sometimes able to generate high-quality completions that incorporate and extend the existing terrain geometry. However, the model fails in roughly $50\%$ of experiments, infilling the remaining portion with terrain that adheres to the biome condition but completely ignores the existing blocks.

We found almost no successful samples when inpainting from human-authored block contexts. While the model occasionally makes trivial extensions of the fixed blocks, it usually completely ignores and/or buries the source block pattern, generating terrain around it. The only plausible result comes from the ``sand donut'' pattern (bottom row, second column), with all other samples failing to incorporate the block pattern in a meaningful way.

\subsection{Sampling Step comparison}

\begin{table}[h]
\centering
\begin{minipage}{0.45\textwidth}
    \centering
    \small
    \label{tab:md4_p2_samplingsteps}
    \caption{Per-biome FID scores for \textbf{MD4} (patch 2), sampled using 1000, 100, and 10 sampling steps. Lower is better. Bold indicates the lowest FID for each biome.}
    \begin{tabular}{lccc}
    \toprule
    Biome & 1000 steps & 100 steps & 10 steps \\
    \midrule
    Beaches & \textbf{63.49} & 89.94 & 193.95 \\
    Birch forest & \textbf{41.83} & 51.93 & 127.01 \\
    Cave & \textbf{46.52} & 55.20 & 77.73 \\
    Desert & \textbf{57.94} & 89.14 & 161.16 \\
    Extreme hills & \textbf{103.78} & 130.05 & 219.18 \\
    Forest & \textbf{72.84} & 75.58 & 137.26 \\
    Ice & \textbf{62.11} & 109.69 & 267.36 \\
    Jungle & \textbf{60.21} & 73.94 & 172.48 \\
    Ocean & \textbf{55.17} & 106.57 & 278.22 \\
    Plains & \textbf{39.31} & 49.94 & 151.79 \\
    River & \textbf{54.81} & 74.30 & 163.72 \\
    Savanna & \textbf{57.60} & 73.42 & 178.24 \\
    Swampland & \textbf{56.92} & 68.77 & 166.95 \\
    Taiga & \textbf{54.71} & 60.67 & 130.63 \\
    Village & \textbf{61.65} & 84.05 & 192.20 \\
    \midrule
    \textbf{Average} & \textbf{59.26} & 79.55 & 174.53 \\
    \bottomrule
    \end{tabular}
\end{minipage}
\hfill
\begin{minipage}{0.45\textwidth}
    \centering
    \small
    \caption{Per-biome FID scores for \textbf{DDPM} (patch 2), sampled using 100, 100, and 10 sampling steps. Lower is better. Bold indicates the lowest FID for each biome.}
    \label{tab:ddpm_samplingsteps}
    \begin{tabular}{lccc}
    \toprule
    Biome & 1000 steps & 100 steps & 10 steps \\
    \midrule
    Beaches & 74.21 & \textbf{65.98} & 74.56 \\
    Birch forest & 45.28 & \textbf{44.15} & 49.92 \\
    Cave & \textbf{55.98} & 57.42 & 77.52 \\
    Desert & 52.42 & \textbf{46.87} & 72.51 \\
    Extreme hills & 86.13 & \textbf{84.26} & 100.57 \\
    Forest & 49.39 & \textbf{47.23} & 55.52 \\
    Ice & \textbf{72.30} & 73.33 & 103.29 \\
    Jungle & \textbf{61.42} & 64.03 & 68.42 \\
    Ocean & 34.66 & 22.26 & \textbf{15.33} \\
    Plains & \textbf{46.02} & 46.58 & 51.70 \\
    River & 59.48 & \textbf{58.11} & 70.04 \\
    Savanna & \textbf{54.32} & 60.15 & 61.14 \\
    Swampland & 81.28 & \textbf{50.29} & 51.97 \\
    Taiga & 57.80 & \textbf{47.56} & 51.34 \\
    Village & \textbf{58.61} & 59.00 & 83.00 \\
    \midrule
    \textbf{Average} & 59.29 & \textbf{55.15} & 65.79 \\
    \bottomrule
    \end{tabular}
\end{minipage}
\end{table}

All visualized samples were generated using the full number of sampling steps ($T = 1000$). This results in approximately 2.5 minutes per $32^3$ chunk for patch 2 models, and 25 seconds for patch 4 models. We investigate the sampling step / quality tradeoff in these models by evaluating samples generated at $1000$, $100$, and $10$ sampling steps. Tables \ref{tab:md4_p2_samplingsteps} and \ref{tab:ddpm_samplingsteps} show the per-biome FID scores for these samples for our MD4 and DDPM patch 2 models.

We observe a sharp degradation in per-biome FID scores for the MD4 models as sample steps decrease. Across all biomes, the lowest per-biome FID scores are achieved at the full 1,000 steps. This pattern does not hold for DDPM, where performance is roughly equivalent between 1,000 and 100 steps, with 100-step samples achieving both a lower average FID, and lower per-biome FID in the majority of biomes. 

We hypothesize that this asymmetry arises from the sampling processes of discrete and continuous diffusion. In MD4, fewer sampling steps means that more tokens must be unmasked simultaneously at each step, with the number of unmasked tokens scaling inversely with sampling steps. Tokens unmasked in the same step are sampled independently --- conditioned on the current partially unmasked state, but blind to each other. As the number of parallelly-sampled tokens grows, so does the risk that these independently plausible predictions combine to create a globally incoherent configuration of blocks. This ``joint distribution issue'', identified by ~\citet{ImprovedVQdiffusion} is inherent to this sampling process.

In our domain, the way this manifests can be easily understood by example. Consider the ocean biome: oceans in Minecraft exist at a fixed ``sea level'', and the surface of the ocean is always flat, at the same height in the vertical axis. If, in an early diffusion step, the model predicts the ocean surface exists at two separate heights (i.e by unmasking air blocks above two water blocks), the diffusion trajectory moves irreparably out of the biome distribution. 

In continuous diffusion, reduced step counts have a different effect. All block embeddings are partially denoised at every sampling step, regardless of the number of sampling steps used. An increase in sampling steps translates to a larger amount of noise being removed from these embeddings at each step. Coarser sampling steps are known to decrease sample quality, introducing a practical trade-off between compute and quality, but this degradation is typically smoother than the effect we observe in MD4.

Several strategies have been proposed to mitigate the joint distribution issue in masked diffusion sampling. \citet{ImprovedVQdiffusion} introduce purity prior sampling, which unmasks tokens where the model assigns the highest probability to its top prediction, reducing the chance that low-confidence tokens are prematurely committed. \citet{kim2025train} propose top-K probability and top-K probability margin inference strategies, which select tokens to unmask based on model confidence and the margin between the top two predictions respectively. Both approaches replace the default random unmasking order with confidence-guided selection, effectively allowing the model to resolve easier tokens first and condition harder decisions on more complete context. We do not experiment with these techniques in this work, but note that they represent a promising avenue for improving reduced-step sample quality, which is an important consideration for deploying these models in interactive or production settings.

\subsection{Biome Block statistics}
We compute two block distribution models on the generated outputs of our models. First, we compute the per-biome vocabulary for all natural biomes, then measure the fraction of that vocabulary that is present in the 100 samples generated for FID computation. Table \ref{tab:voxel_coverage_non_air} shows these results. Between our five models, the MD4 patch 4 village boosted model performs the best, with the highest coverage of the true block vocabulary 12 of 15 biomes.  The natural occurrence model achieves significantly higher vocabulary coverage in the cave and village biomes, the two most underrepresented biomes present in its dataset. 

We also compute the Jensen-Shannon divergence between the block distributions of generated vs real samples, shown in table \ref{tab:voxel_js_non_air}. The MD4 patch 2 model performs best on average, and we see similar trends as in the per-biome FID scores: village boosted has low divergence in the village biome, while natural occurrence has lowest divergence in its highest represented biomes of ocean.

Overall, these metrics represent very coarse aggregate measures of distribution fit, and do not directly measure sample quality. We do not observe meaningful correlations between performance in these metrics and qualitative sample quality. While we find some alignment intuitive alignment (particularly in the village biome), these simple metrics do not fully address the need for evaluation of quality in 3D domains.

\begin{table*}[h]
\centering
\caption{Per-biome non-air vocabulary coverage for each model. Coverage is the fraction of block types present in real chunks that also appear in generated samples. Higher is better.}
\label{tab:voxel_coverage_non_air}
\begin{tabular}{lccccc}
\toprule
Biome & MD4 p2 & DDPM p2 & MD4 p4 Balanced & MD4 p4 Natural & MD4 p4 Boosted \\
\midrule
beaches & 0.640 & 0.291 & 0.500 & 0.302 & \textbf{0.709} \\
birch\_forest & 0.451 & 0.256 & 0.390 & 0.415 & \textbf{0.549} \\
cave & 0.460 & 0.356 & 0.644 & \textbf{0.908} & 0.563 \\
desert & 0.382 & 0.211 & 0.421 & 0.632 & \textbf{0.816} \\
extreme\_hills & 0.346 & 0.244 & 0.449 & 0.410 & \textbf{0.538} \\
forest & 0.337 & 0.232 & 0.411 & 0.389 & \textbf{0.463} \\
ice & 0.470 & 0.289 & 0.470 & 0.386 & \textbf{0.566} \\
jungle & 0.506 & 0.289 & 0.422 & 0.506 & \textbf{0.530} \\
ocean & \textbf{0.945} & 0.151 & 0.753 & 0.192 & 0.877 \\
plains & 0.407 & 0.259 & 0.420 & 0.543 & \textbf{0.568} \\
river & 0.439 & 0.347 & 0.582 & 0.551 & \textbf{0.939} \\
savanna & 0.354 & 0.266 & 0.544 & 0.392 & \textbf{0.557} \\
swampland & 0.534 & 0.342 & 0.603 & 0.521 & \textbf{0.973} \\
taiga & 0.388 & 0.300 & 0.487 & 0.425 & \textbf{0.525} \\
village & 0.880 & 0.733 & 0.880 & \textbf{0.973} & 0.920 \\
\midrule
\textbf{Average} & 0.503 & 0.304 & 0.532 & 0.503 & \textbf{0.673} \\
\bottomrule
\end{tabular}
\end{table*}

\begin{table*}[h]
\centering
\caption{Per-biome non-air Jensen--Shannon divergence between generated and real block-name distributions. Lower is better; zero indicates an exact match.}
\label{tab:voxel_js_non_air}
\begin{tabular}{lccccc}
\toprule
Biome & MD4 p2 & DDPM p2 & MD4 p4 Balanced & MD4 p4 Natural & MD4 p4 Boosted \\
\midrule
beaches & 0.004 & 0.056 & \textbf{0.002} & 0.005 & 0.003 \\
birch\_forest & 0.006 & 0.009 & \textbf{0.003} & 0.008 & 0.006 \\
cave & \textbf{0.010} & 0.010 & 0.015 & 0.032 & 0.018 \\
desert & \textbf{0.013} & 0.057 & 0.026 & 0.017 & 0.021 \\
extreme\_hills & 0.012 & 0.019 & 0.006 & \textbf{0.005} & 0.010 \\
forest & 0.044 & \textbf{0.017} & 0.047 & 0.040 & 0.046 \\
ice & 0.004 & 0.015 & \textbf{0.002} & 0.009 & 0.005 \\
jungle & 0.023 & 0.030 & 0.014 & \textbf{0.012} & 0.016 \\
ocean & 0.012 & 0.036 & 0.008 & \textbf{0.003} & 0.008 \\
plains & 0.004 & 0.009 & 0.003 & 0.005 & \textbf{0.003} \\
river & \textbf{0.004} & 0.015 & 0.019 & 0.015 & 0.067 \\
savanna & \textbf{0.004} & 0.010 & 0.005 & 0.006 & 0.005 \\
swampland & 0.005 & 0.016 & 0.005 & \textbf{0.001} & 0.012 \\
taiga & 0.013 & 0.019 & \textbf{0.010} & 0.014 & 0.011 \\
village & 0.011 & 0.008 & 0.007 & 0.022 & \textbf{0.002} \\
\midrule
\textbf{Average} & \textbf{0.011} & 0.022 & 0.012 & 0.013 & 0.016 \\
\bottomrule
\end{tabular}
\end{table*}

\subsection{Preference Study details}
\label{sec:apdx_user_study}

\paragraph{Study design and goals}
We design our preference study to investigate two research questions: 
(i) Are generated Minecraft chunks comparable in quality to real Minecraft terrain? 
(ii) Is render-based FID aligned with human preference?

We use a a 2 alternative force-choice format, where participants compare terrain from two sources and select the better one. We display four chunks from each source to better represent the quality of the source, rather than the quality of a single chunk at a time. We include four sources of data: MD4 (patch size 2), MD4 (patch size 4), DDPM (patch size 2), and real chunks samples from the balanced validation set. Each source contains 100 samples for each natural biome. For the generative models, these are the same samples used to compute the per-biome FID scores reported in section \ref{sec4:experiments}. During each trial, one of six possible pairings are selected, and ordering (left vs. right) is randomized. A biome is randomly selected, and four samples are drawn from each source randomly without replacement. This allows us to study both real vs. generated preferences, as well as model vs model preferences.

\paragraph{Data collection}
We recruit 19 participants from our institution, and all express a familiarity with Minecraft and the biome system. Each participant is asked to complete 60 trials, and all participants complete the full study. Participants are asked to provide a participant ID, which may be a pseudonym, and not other data is collected other than the selections they make. All users are shown the following instructions before beginning the study:

\begin{quote}
In each trial, you will see two candidate sets of Minecraft terrain, labeled 1 and 2.
Each candidate contains four terrain chunks. Both candidates are intended to represent the same target biome, which will be shown at the top of the screen.

Your task is to choose the better overall set. You will be asked to complete 60 trials as part of this study.

When making your choice, use the following priority:

1. Visual quality and match to the target biome
   Prefer the set that looks more like plausible Minecraft terrain for the given biome and has fewer obvious artifacts, unnatural structures, broken patterns, or other visual errors.
2. Overall preference if still tied in your mind
   If the two sets seem equally good with respect to both quality and biome match, choose the one you personally prefer overall.

Terrain chunks may be generated by AI systems, or may come directly from the game. Please judge only what you see on the screen. Do not try to guess where the terrain came from. There is no tie option, so please choose the set you think is better overall, even if the difference is small.

You can zoom, rotate, and pan to see the terrain chunks from all angles using the mouse. Click and drag to rotate, scroll to zoom, and hold middle mouse button to pan.

We do not collect personally identifiable information besides a participant ID as part of this study. Your responses will be recorded and may be analyzed in aggregate for research purposes, including publications or presentations. No information that directly identifies you will be included in any paper or public release.
\end{quote}

\begin{figure}[!h]
    \centering
    \includegraphics[width=0.6\linewidth]{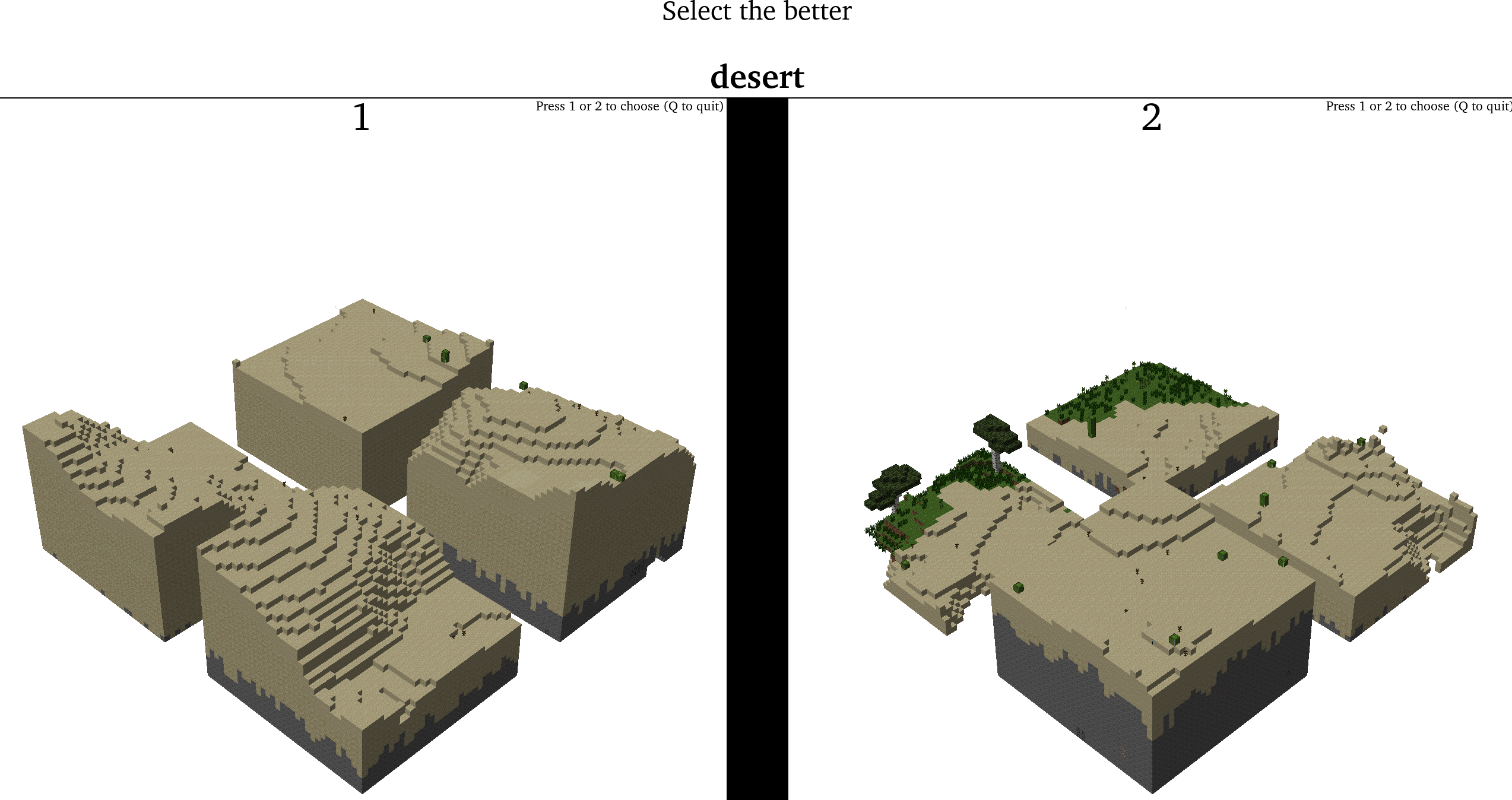}
    \caption{Example of the user study interface. Each candidate set is rendered in 3D using the in-game Minecraft textures, and can be rotated independently to view chunks from all angles.}
    \label{fig:user_study_screenshot}
\end{figure}

\paragraph{Interface}
Figure \ref{fig:user_study_screenshot} shows an example of the user study interface, comparing two samples for the desert biome.

\paragraph{Analysis}
We aggregate the data from all user trials, anonymizing users. Aggregating results by model presence, we compute a win rate (number of trials preferred / number of trials present) in Table \ref{tab:user-study-overall-model-stats}. We also construct a pairwise win-rate matrix for each pairing of sources (table \ref{tab:user-study-pairwise-win-rates}), showing how each source performed against all others. We finally compute per-biome win rates for each model, reported in table \ref{tab:user-study-per-biome}.

We report 95\% Wilson confidence intervals and number of trials for tables \ref{tab:user-study-overall-model-stats} and \ref{tab:user-study-pairwise-win-rates}. We also conduct a one-sided binomial test under the null hypothesis that data source preference is chance (50\%).

For alignment with FID, we consider only trials between model sources, and exclude real data, leaving $512$ trials. For each remaining trial, we compute the absolute difference between the per-biome FID scores for the biome shown. We use the per-biome FID scores shown in Figure \ref{fig:md4_vs_md4_vs_ddpm}, treating them as representative of the four randomly selected samples shown. We then compute the percentage of trials where the participant chose the lower-FID source, and compute a similar one-sided binomial test against random chance (Table \ref{tab:fid-alignment}, row 1).

We conduct the same analysis, discarding trials where the absolute FID difference was below a certain threshold (Table \ref{tab:fid-alignment}, rows 2-4). We also analyze the alignment between FID and preference on a per-biome basis (Table \ref{tab:fid-alignment-biome}). On a per-biome basis, agreement was highest in the village and desert biomes, while all other biomes have confidence intervals overlapping 0.5.

\begin{table}[h]
\caption{Preference-FID model by biome on model-vs-model trials. Agreement is the proportion of trials where which the participant selected the model with the lower per-biome FID. Confidence intervals are Wilson 95\% intervals.}
\label{tab:fid-alignment-biome}
\small
\centering
\begin{tabular}{llll}
\toprule
Biome & Agreement & $n$ & 95\% CI \\
\midrule
beaches & 47.5\% & 40 & [32.9\%, 62.5\%] \\
birch\_forest & 43.8\% & 32 & [28.2\%, 60.7\%] \\
cave & 42.9\% & 35 & [28.0\%, 59.1\%] \\
desert & 70.0\% & 30 & [52.1\%, 83.3\%] \\
extreme\_hills & 46.4\% & 28 & [29.5\%, 64.2\%] \\
forest & 50.0\% & 40 & [35.2\%, 64.8\%] \\
ice & 59.5\% & 37 & [43.5\%, 73.7\%] \\
jungle & 41.4\% & 29 & [25.5\%, 59.3\%] \\
ocean & 59.3\% & 27 & [40.7\%, 75.5\%] \\
plains & 50.0\% & 38 & [34.8\%, 65.2\%] \\
river & 63.3\% & 49 & [49.3\%, 75.3\%] \\
savanna & 54.5\% & 33 & [38.0\%, 70.2\%] \\
swampland & 53.6\% & 28 & [35.8\%, 70.5\%] \\
taiga & 51.9\% & 27 & [34.0\%, 69.3\%] \\
village & 74.4\% & 39 & [58.9\%, 85.4\%] \\
\bottomrule
\end{tabular}
\end{table}

\begin{table}[h]
\caption{Overall participant preference by model.}
\label{tab:user-study-overall-model-stats}
\centering
\small
\begin{tabular}{lrrl}
\toprule
Model & Wins & Appearances & Preference rate (95\% CI) \\
\midrule
MD4 p2 & 290 & 505 & 57.4\% [53.1, 61.7] \\
DDPM p2 & 282 & 530 & 53.2\% [49.0, 57.4] \\
MD4 p4 & 255 & 517 & 49.3\% [45.0, 53.6] \\
Real & 213 & 528 & 40.3\% [36.2, 44.6] \\
\bottomrule
\end{tabular}
\end{table}

\begin{table}[h]
\caption{Head-to-head preference matrix between data sources. Each cell gives P(row model preferred over column model) with matchup count n.}
\label{tab:user-study-pairwise-win-rates}
\centering
\small
\begin{tabular}{lllll}
\toprule
Model & MD4 p2 & DDPM p2 & MD4 p4 & real\_samples \\
\midrule
MD4 p2 & -- & 49.4\% (n=158) & 55.2\% (n=174) & 67.1\% (n=173) \\
DDPM p2 & 50.6\% (n=158) & -- & 53.3\% (n=180) & 55.2\% (n=192) \\
MD4 p4 & 44.8\% (n=174) & 46.7\% (n=180) & -- & 57.1\% (n=163) \\
real\_samples & 32.9\% (n=173) & 44.8\% (n=192) & 42.9\% (n=163) & -- \\
\bottomrule
\end{tabular}
\end{table}

\begin{table}[h]
\caption{Per-biome normalized preference rates (wins / appearances within each biome). Bold indicates the highest source in each biome. Total n gives the total number of source appearances in that biome across all source match-ups.}
\label{tab:user-study-per-biome}
\centering
\small
\begin{tabular}{lrllll}
\toprule
Biome  & MD4 p2 & DDPM p2 & MD4 p4 & Real \\
\midrule
beaches  & \textbf{68.4\%} & 47.2\% & 34.3\% & 48.3\% \\
birch\_forest  & 54.5\% & \textbf{59.4\%} & 53.6\% & 31.0\% \\
cave  & \textbf{57.1\%} & 52.8\% & 54.9\% & 38.5\% \\
desert  & 48.5\% & \textbf{80.6\%} & 34.5\% & 36.4\% \\
extreme\_hills  & \textbf{63.3\%} & 50.0\% & 61.8\% & 27.8\% \\
forest  & 42.1\% & 53.8\% & \textbf{57.1\%} & 46.9\% \\
ice  & 57.6\% & 30.6\% & 48.7\% & \textbf{64.7\%} \\
jungle  & 61.0\% & \textbf{72.4\%} & 47.2\% & 29.2\% \\
ocean  & 24.0\% & \textbf{72.0\%} & 60.7\% & 41.7\% \\
plains  & \textbf{59.1\%} & 55.8\% & 55.6\% & 29.4\% \\
river  & \textbf{65.3\%} & 30.6\% & 52.5\% & 52.5\% \\
savanna  & \textbf{70.0\%} & 43.2\% & 40.0\% & 50.0\% \\
swampland  & \textbf{58.6\%} & 36.7\% & 51.6\% & 52.9\% \\
taiga  & 50.0\% & 66.7\% & \textbf{69.0\%} & 21.4\% \\
village  & \textbf{68.3\%} & 64.5\% & 16.1\% & 44.0\% \\
\bottomrule
\end{tabular}
\end{table}

\clearpage
\subsection{Training Configuration}
\begin{figure}[t]
    \centering
    \begin{minipage}{0.48\linewidth}
        \centering
        \includegraphics[width=\linewidth]{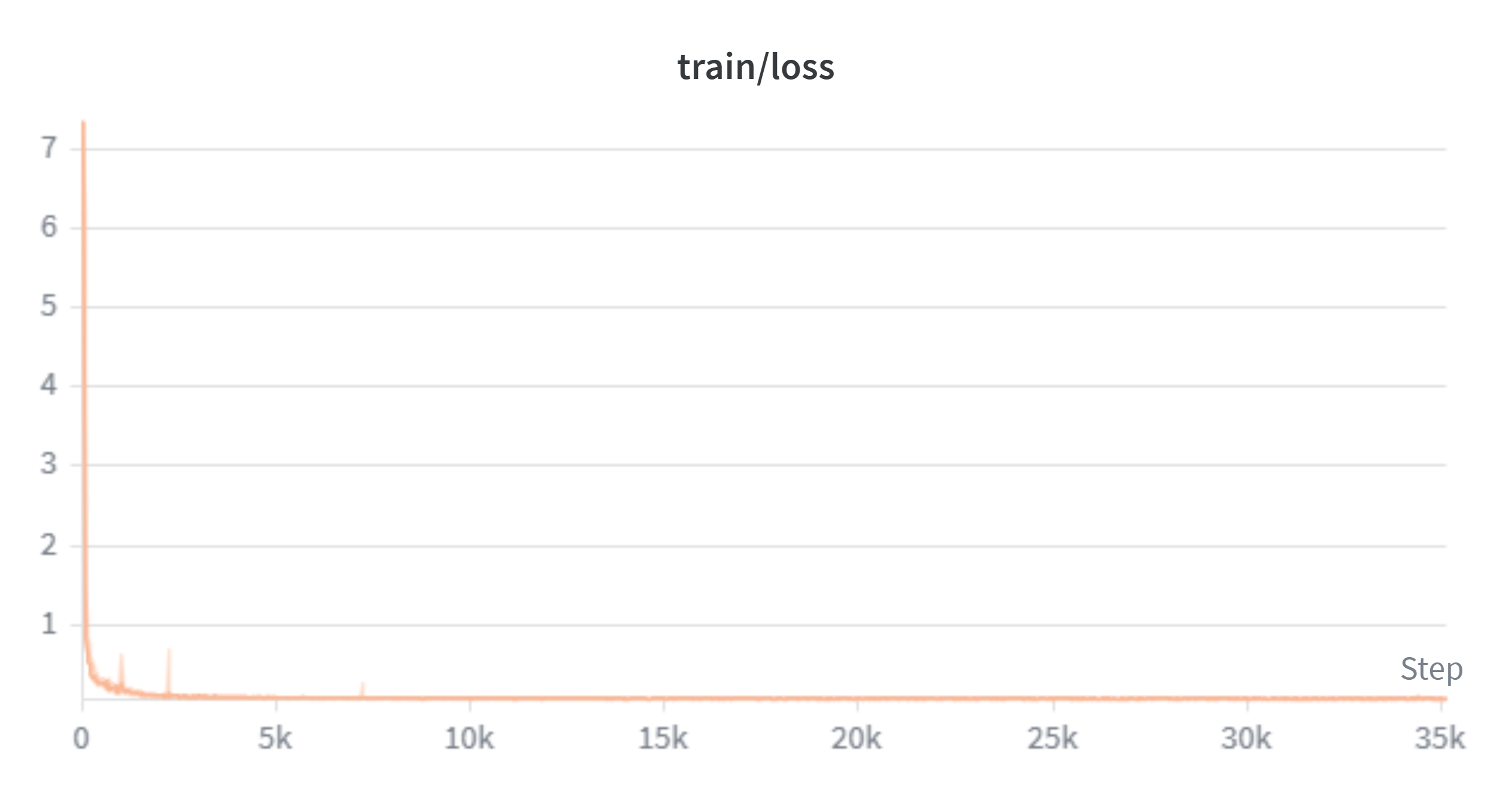}
    \end{minipage}
    \hfill
    \begin{minipage}{0.48\linewidth}
        \centering
        \includegraphics[width=\linewidth]{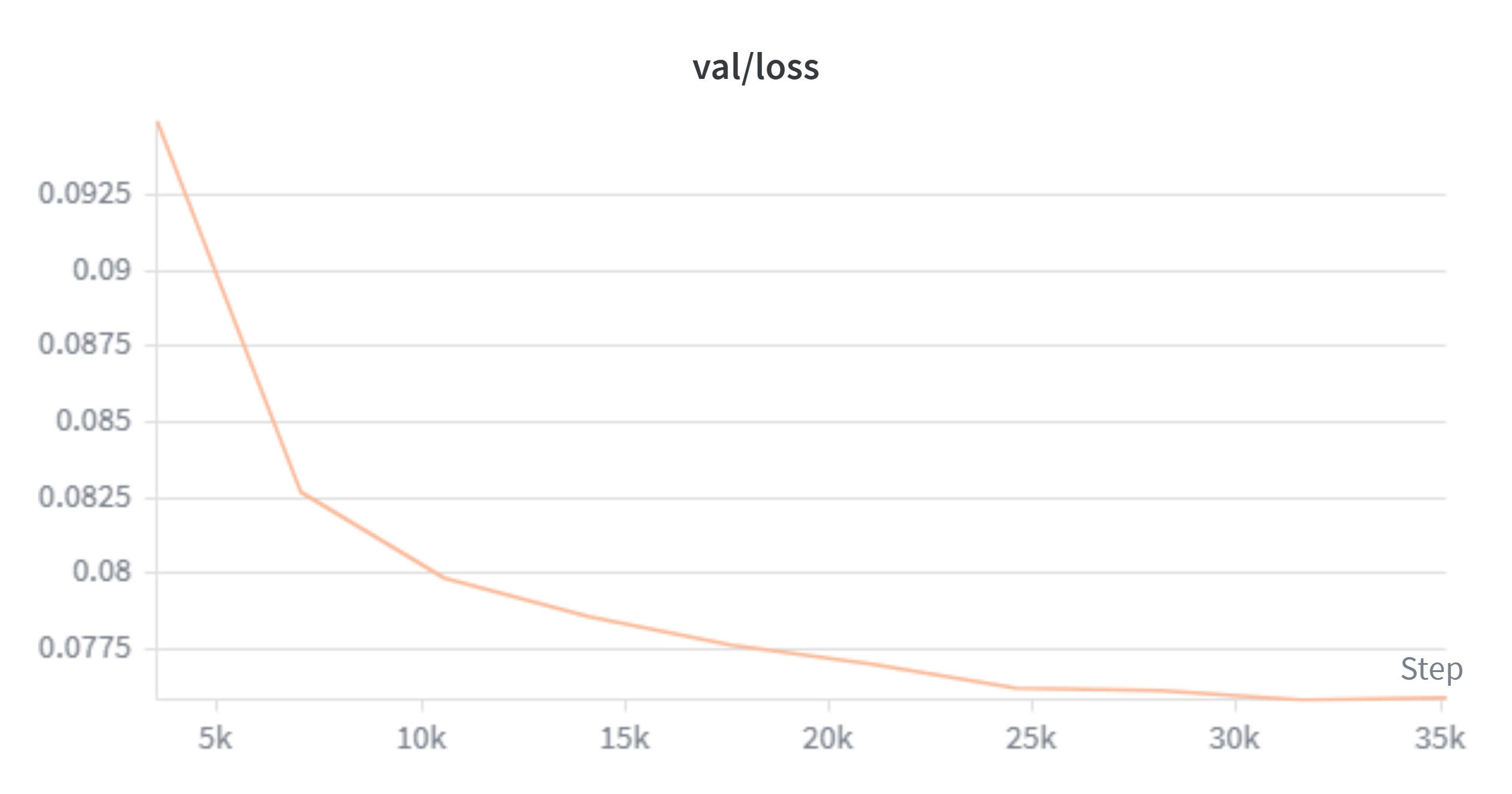}
    \end{minipage}
    \caption{Train and validation loss curves for MD4 patch 2 model}
    \label{fig:md4p2_loss}
    \centering
    \begin{minipage}{0.48\linewidth}
        \centering
        \includegraphics[width=\linewidth]{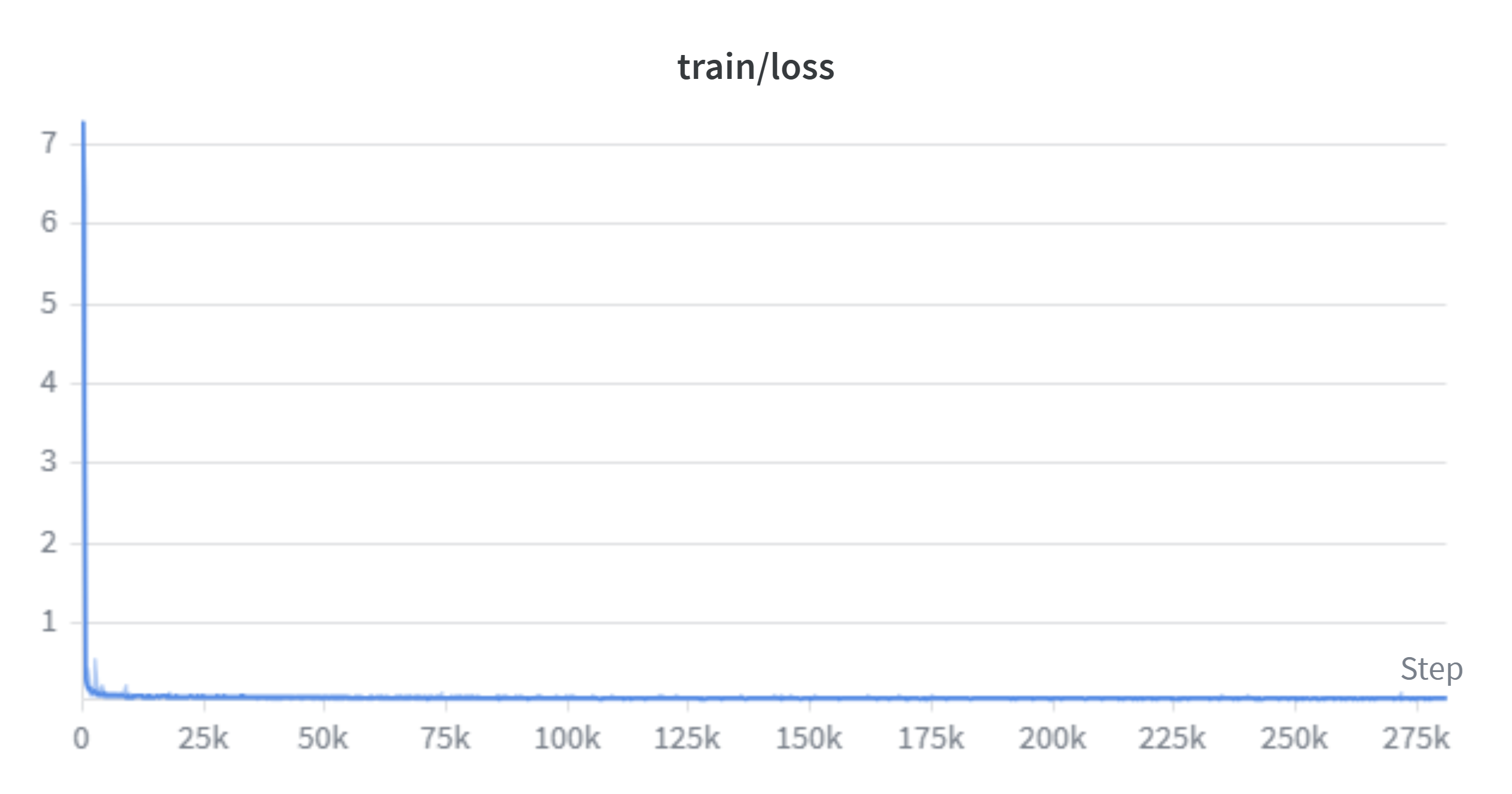}
    \end{minipage}
    \hfill
    \begin{minipage}{0.48\linewidth}
        \centering
        \includegraphics[width=\linewidth]{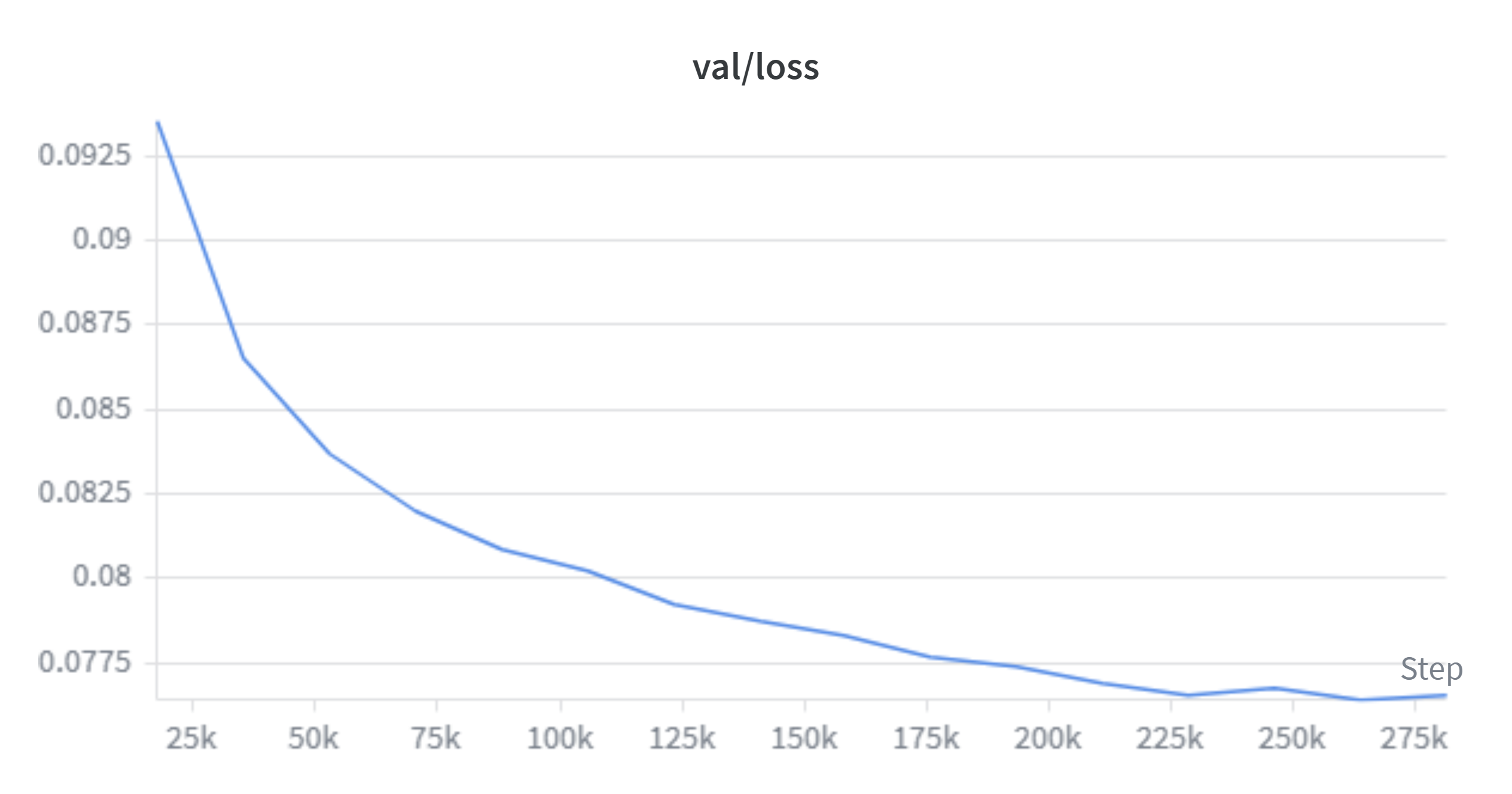}
    \end{minipage}
    \caption{Train and validation loss curves for MD4 patch 4 model}
    \label{fig:md4p4_loss}
    \centering
    \begin{minipage}{0.48\linewidth}
        \centering
        \includegraphics[width=\linewidth]{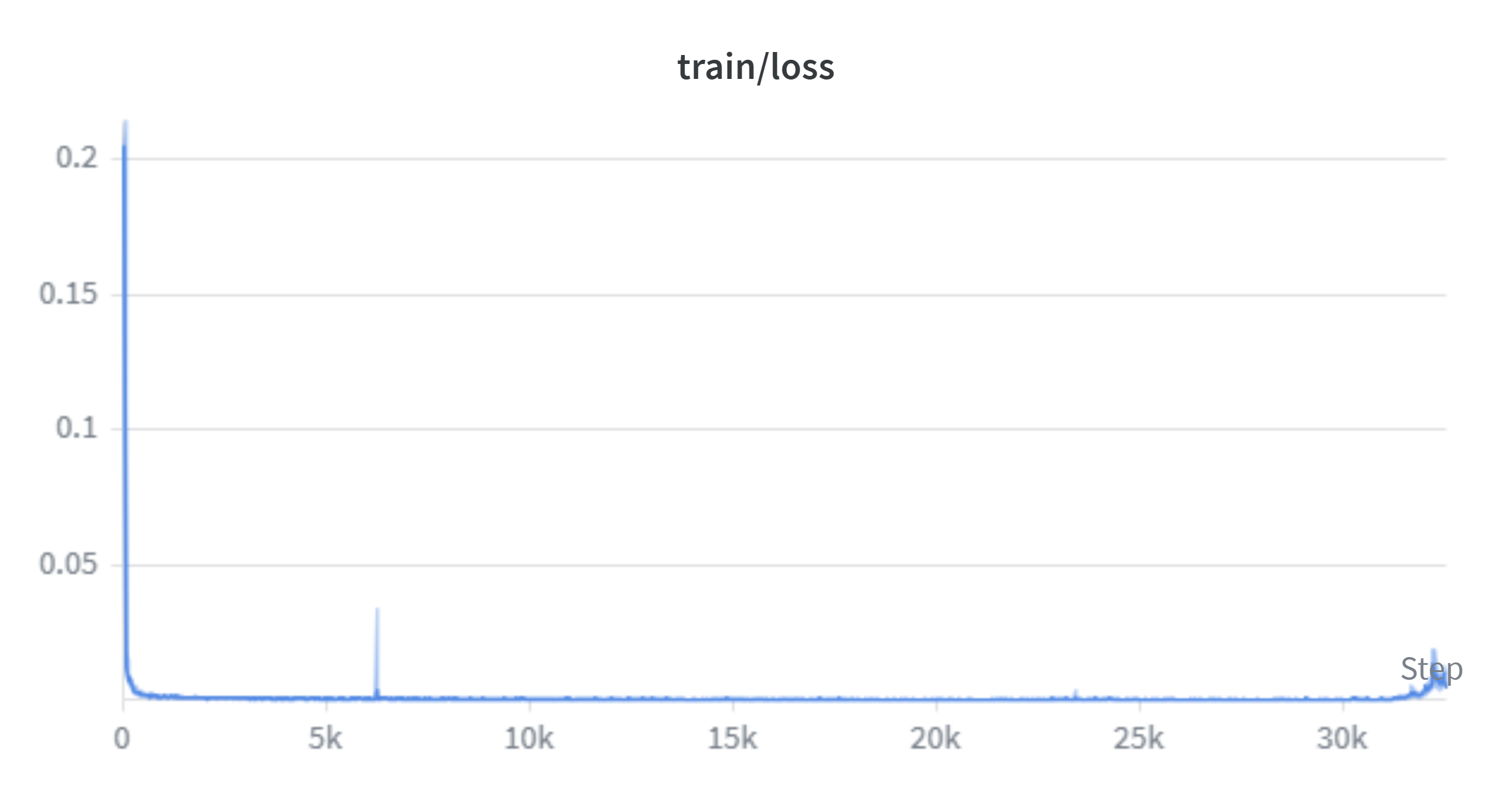}
    \end{minipage}
    \hfill
    \begin{minipage}{0.48\linewidth}
        \centering
        \includegraphics[width=\linewidth]{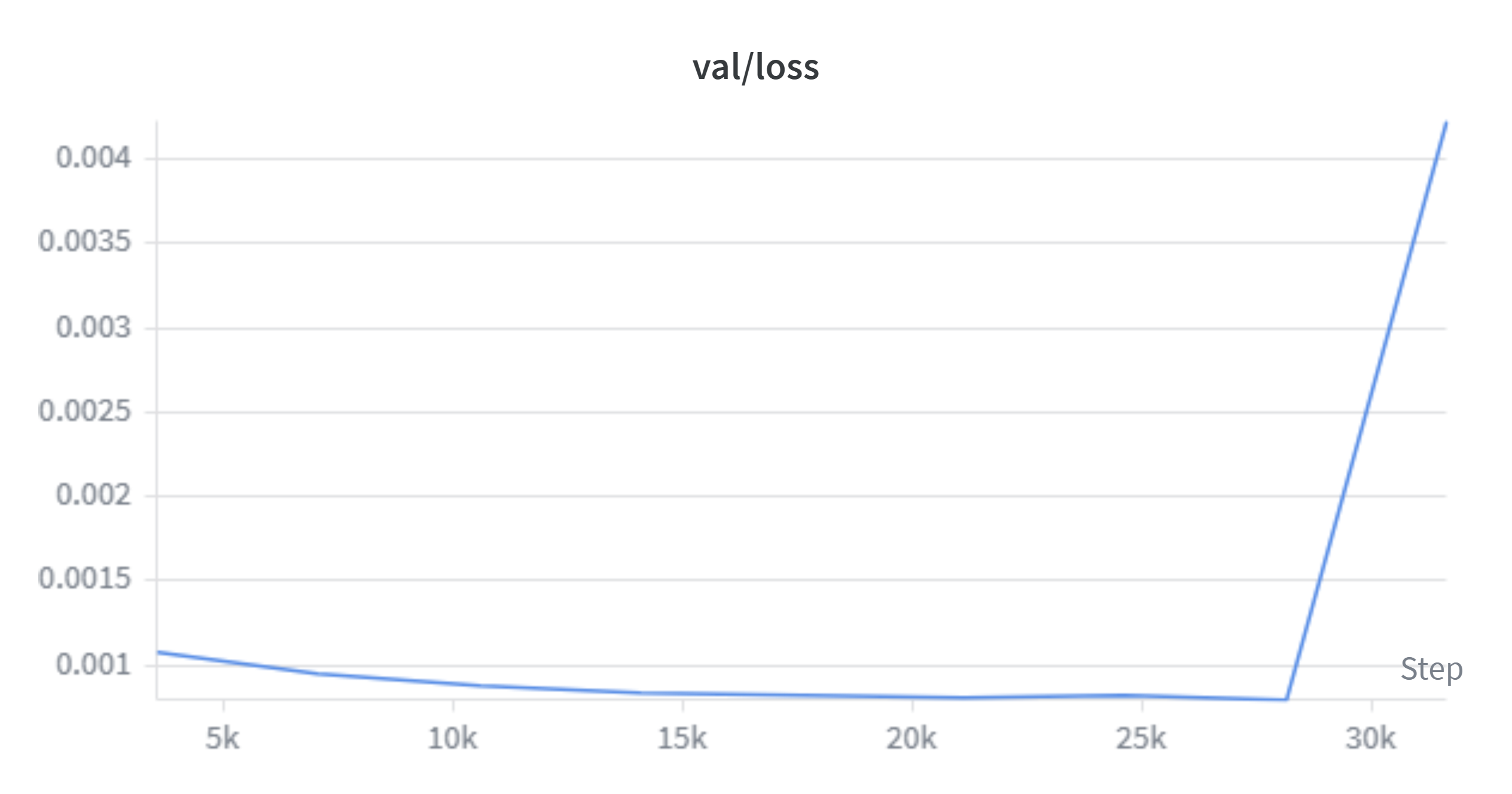}
    \end{minipage}
    \caption{Train and validation loss curves for DDPM patch 2 model}
    \label{fig:ddpmp2_loss}
\end{figure}

Both diffusion formulations use classifier-free guidance (CFG) for conditional sample generation. During training, we randomly drop the biome condition with probability $p_\text{drop} = 0.2$, and at sampling time we apply guidance with scale for conditional generation $4.0$. All models incorporate random 90-degree rotation about the vertical axis as a data augmentation during training. We train all models using the AdamW optimizer, with a learning rate of $3e^{-4}$ for patch 4 models and $1e^{-3}$ for patch 2 models, with weight decay $=0.01$ \cite{loshchilov2019decoupled}. We select these parameters via hyperparameter sweep, optimizing for validation loss. We use an effective batch size of $512$ distributed across 4 H100 GPUs. Patch 2 models are trained for 20 epochs, and patch 4 models trained for 160 epochs, equating the number of patch-tokens seen during training. Total wall-clock training time is approximately $192$ GPU-hours for all models. We show the training and validation loss curves for our MD4 (patch 2), MD4 (patch 4), and DDPM (patch 4) models in Figure \ref{fig:md4p2_loss}.

\onecolumn

\end{document}